\documentclass{article}

\usepackage[final]{neurips_2024}

\usepackage{amsmath}
\usepackage{xcolor}

\usepackage[utf8]{inputenc} 
\usepackage[T1]{fontenc}    
\usepackage{hyperref}       
\usepackage{cleveref}       
\usepackage{url}            
\usepackage{booktabs}       
\usepackage{amsfonts}       
\usepackage{nicefrac}       
\usepackage{microtype}      
\usepackage{xcolor}         
\usepackage{graphicx} 
\usepackage{subcaption}

\usepackage{algorithm}
\usepackage{algpseudocode}
\usepackage{algorithmicx}
\usepackage{caption}

\newcommand{\intellect}{\textsc{intellect-2}}
\newcommand{\primeframework}{\textsc{prime-rl}}
\newcommand{\shortexperiment}{\textsc{target-short}}
\newcommand{\longexperiment}{\textsc{target-long}}
\newcommand{\genesys}{\textsc{genesys}}
\newcommand{\toploc}{\textsc{toploc}}
\newcommand{\shardcast}{\textsc{shardcast}}

\title{INTELLECT-2: A Reasoning Model Trained Through Globally Decentralized Reinforcement Learning}

\author{%
  Prime Intellect Team \\
\AND
  Sami Jaghouar \\
\And
  Justus Mattern \\
\And
  Jack Min Ong \\
\And
  Jannik Straube \\  
\AND
  Manveer Basra \\
\And
Aaron Pazdera \\
\And
Kushal Thaman \\
\And
Matthew Di Ferrante \\
\And
Felix Gabriel \\
\And
  Fares Obeid \\
\And
Kemal Erdem \\
\And
Michael Keiblinger \\
\And
  Johannes Hagemann \\
}

\begin{document}

{
\begingroup
\begin{figure*}
    \centering
    \includegraphics[width=0.125\textwidth]{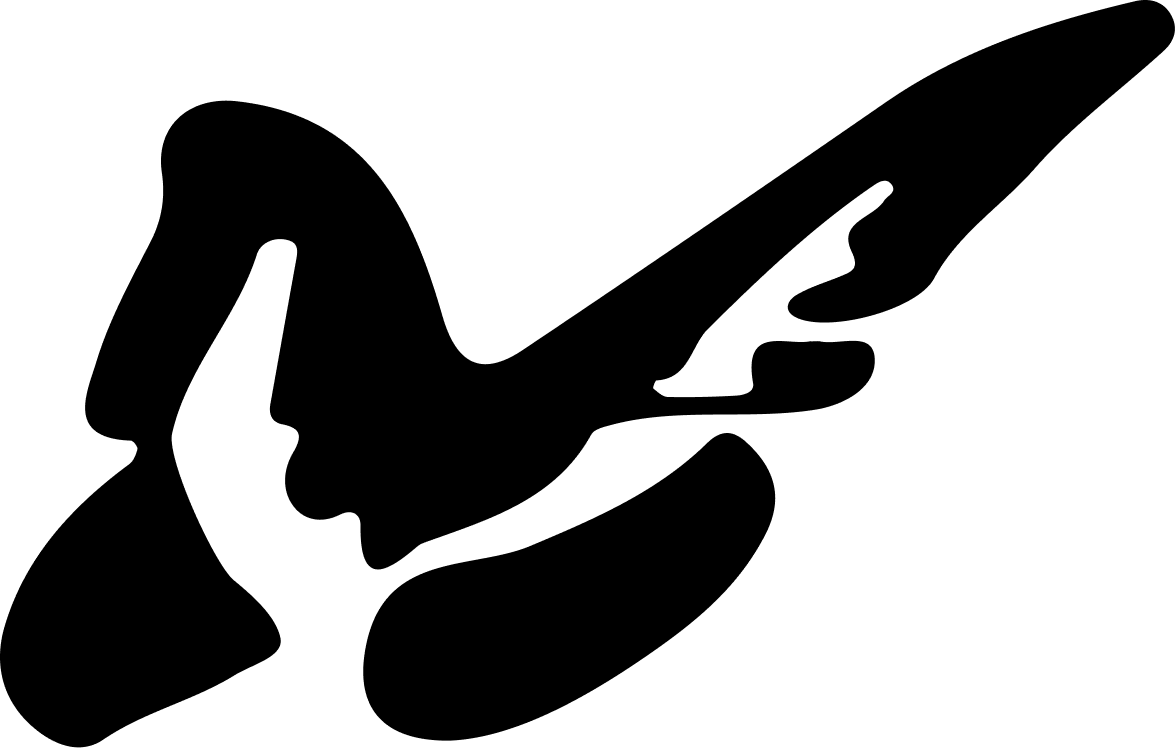}
\end{figure*}
\endgroup
}
\setcounter{figure}{0}

\maketitle

\begin{abstract}
We introduce \intellect, the first globally distributed reinforcement learning (RL) training run of a 32 billion parameter language model.
Unlike traditional centralized training efforts, \intellect \ trains a reasoning model using fully asynchronous RL across a dynamic, heterogeneous swarm of permissionless compute contributors.

To enable a training run with this unique infrastructure, we built various components from scratch: we introduce \primeframework, our training framework purpose-built for distributed asynchronous reinforcement learning, based on top of novel components such as \toploc, which verifies rollouts from untrusted inference workers, and \shardcast, which efficiently broadcasts policy weights from training nodes to inference workers.

Beyond infrastructure components, we propose modifications to the standard GRPO training recipe and data filtering techniques that were crucial to achieve training stability and ensure that our model successfully learned its training objective, thus improving upon QwQ-32B, the state of the art reasoning model in the 32B parameter range.

We open-source \intellect \ along with all of our code and data, hoping to encourage and enable more open research in the field of decentralized training.
\end{abstract}

\newpage

\tableofcontents

\newpage

\section{Introduction}

Test-time compute scaling with reinforcement learning has emerged as a new scaling axis for large language models (LLMs), enabling improvements by allowing models to spend more time reasoning.
The success of RL approaches used in models like DeepSeek-R1 \citep{deepseekai2025deepseekr1incentivizingreasoningcapability}, QwQ \citep{qwq32b}, and OpenAI’s o1 \citep{openai2024openaio1card} in learning reasoning capabilities highlight the power of test-time compute scaling.

However, reinforcement learning training is typically centralized, requiring large clusters of co-located GPUs and fast interconnect speeds. With \intellect, we showcase a paradigm shift: reinforcement learning is inherently more asynchronous and well suited for decentralized, globally distributed compute.

In this paper, we present the first large-scale experiment to collaboratively train a 32-billion-parameter language model using reinforcement learning across a permissionless, globally distributed network of contributors.

We open-source the \intellect\ model, tasks and verifier environments at \href{https://huggingface.co/PrimeIntellect/INTELLECT-2}{\texttt{huggingface.co/PrimeIntellect/INTELLECT-2}} and the \primeframework\ framework for globally distributed RL training at \href{https://github.com/PrimeIntellect-ai/prime-rl}{\texttt{github.com/PrimeIntellect-ai/prime-rl}}.

The remainder of this report is organized as follows: \Cref{sec:infra} provides a detailed overview of the decentralized training infrastructure including \primeframework, \toploc, \shardcast\ and the compute orchestration protocol. \Cref{sec:training-recipe} describes the RL training recipe used to train \intellect. \Cref{sec:experiment} presents the experiments we ran along with training performance metrics and model evaluation results. \Cref{sec:discussion-decentralized-ai} discusses possible implications of decentralized training in the test-time compute paradigm. Finally, \Cref{sec:conclusion} concludes the report and outlines directions for future work.

\section{Training Infrastructure}
\label{sec:infra}

We introduce the following key open-source infrastructure components for training \intellect:

\begin{itemize}
    \item \textbf{\primeframework}: A fully asynchronous reinforcement learning framework designed for decentralized training. The decoupling of rollout generation, model training, and weight broadcasting, enables training across heterogeneous, unreliable networks.

    \item \textbf{\shardcast:} A library for distributing large files via a HTTP-based tree-topology network that efficiently propagates updated model weights to the decentralized inference workers.
    
    \item \textbf{\toploc~\citep{toploc}:} A locality-sensitive hashing scheme for efficient verifiable inference. It detects tampering or precision changes in model inference and works reliably across non-deterministic GPU hardware.

    \item \textbf{Protocol Testnet:} Provides the infrastructure to aggregate and coordinate global compute resources.
    
\end{itemize}

\begin{figure*}
  \centering
  \includegraphics[width=\linewidth]{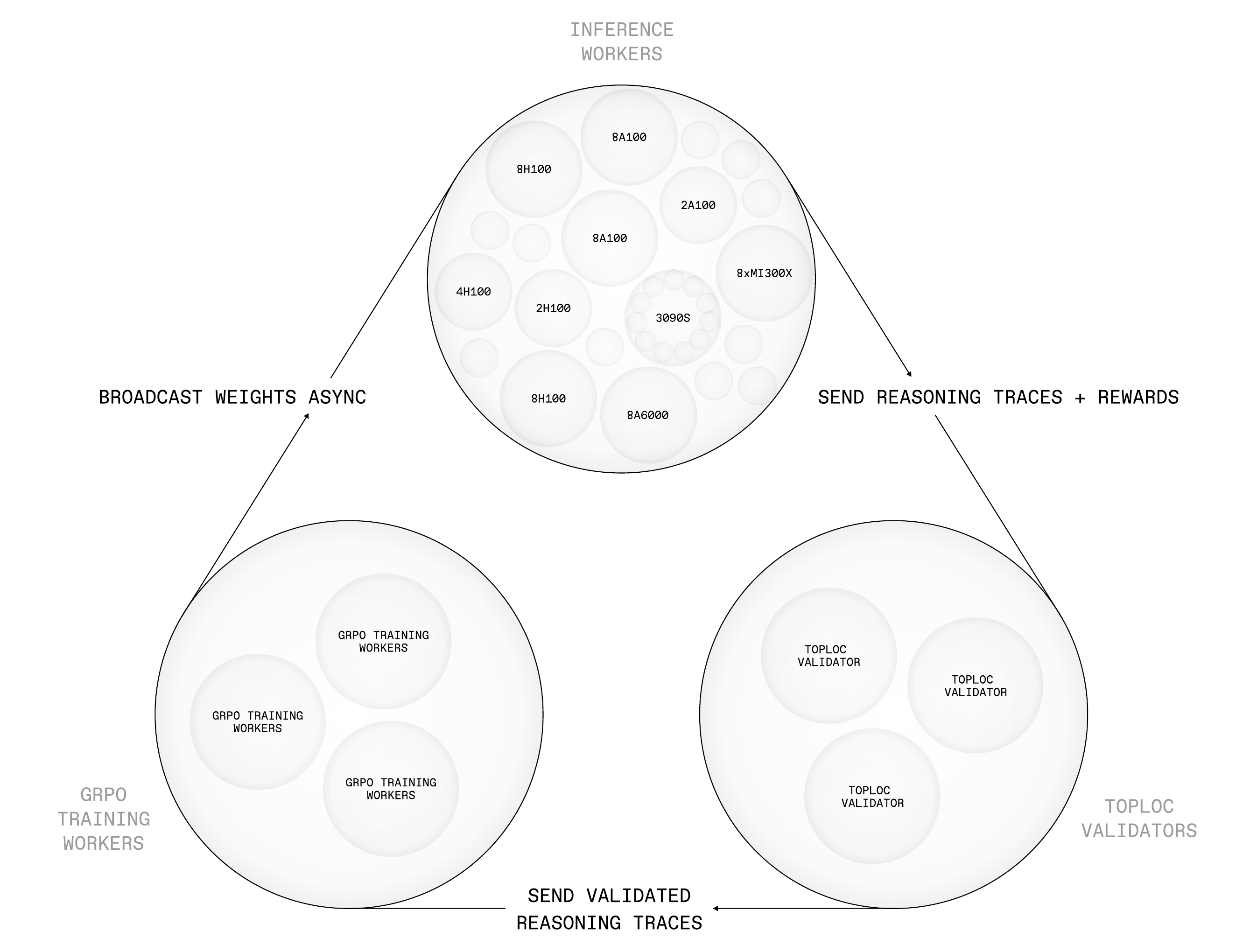}
  \caption{System Overview of \intellect\ Distributed RL Training Infrastructure.}
  \label{fig:system-overview}
\end{figure*}

 As shown in \autoref{fig:system-overview}, the \intellect\ infrastructure, built using these components, is structured around three primary roles: Inference rollout workers that generate reasoning traces using the current policy; \toploc\ validators that verify the integrity of these rollouts; and GRPO training workers that aggregate verified data, update the policy using the GRPO algorithm, and distribute new weights via \shardcast.

This decentralized RL training setup offers several key advantages:

\begin{itemize}
    \item \textbf{No communication overhead:} By leveraging asynchronous reinforcement learning, the broadcast of new policy weights is fully overlapped with ongoing inference and training—eliminating the communication bottleneck.

    \item \textbf{Support for heterogeneous nodes:} Contributors can generate rollouts at their own pace using various hardware; there is no requirement for uniform speed across nodes.

    \item \textbf{Low resource requirements:} Inference workers, which constitute the majority of compute in this setup, can run on consumer-grade GPUs.

    \item \textbf{Efficient validation:} \toploc\ performs validation significantly faster than generation.
\end{itemize}

\subsection{\primeframework: A Framework for Distributed Asynchronous Reinforcement Learning}

We developed a new framework, \primeframework\footnote{https://github.com/PrimeIntellect-ai/prime-rl}, to support training and inference workloads for reinforcement learning.
Unlike existing frameworks such as verl \cite{sheng2024hybridflow} and TRL \cite{vonwerra2022trl}, which execute training and inference sequentially within the same process, \primeframework\ natively enables asynchronous execution of training and inference.
This decoupling allows model updates to be computed on trusted centralized nodes, while rollouts are independently generated on trustless decentralized nodes.

\primeframework's architecture completely separates training and inference components into distinct executable files that communicate only when exchanging data and checkpoints.
This clean separation eliminates the need for centralized orchestrators like Ray \citep{moritz2018raydistributedframeworkemerging}, and our two-step asynchronous design effectively hides latency that would typically be associated with data transfers, creating an efficient distributed reinforcement learning pipeline.

\subsubsection{Training}

To reduce GPU memory requirements during training, we shard the model weights, gradients, and optimizer states across GPUs using PyTorch FSDP2 \cite{zhao2023pytorchfsdpexperiencesscaling}, following a strategy similar to ZeRO-3 \cite{rajbhandari2020zeromemoryoptimizationstraining}.
We load training data from remote storage, and rollout data is exchanged between inference workers and the trainer using Parquet files.

The asynchronous nature of rollout generation is transparent to the trainer, as we compute log-probabilities using the policy at the start of the optimization step rather than the policy that produced the original trajectories.
This design choice aligns with the implementations in verl~\cite{sheng2024hybridflow} which we used as a reference.

In its current form, \primeframework\ implements GRPO training, along with auxiliary KL and entropy losses.

\subsubsection{Inference}

To generate the rollouts, we use vLLM \cite{kwon2023vllm}, loading the model in bfloat16 precision.
For each batch, input questions are sampled randomly using a deterministic seed to prevent inference workers from selecting easy samples, as detailed in Section~\ref{fixed_data_sampling}.

To support \toploc \ proof construction, we capture the final hidden states using a hook in the logits processor.
Instead of storing the activations for the full sequence, we incrementally store and hash them at every 32-token interval which reduces the memory overhead of the proof construction.
To reduce blocking overhead, the proof construction is performed asynchronously on the CPU in parallel with the GPU forward pass.
Together, these optimizations limit the overhead of proof generation to only a $\sim 1\%$ reduction in tokens-per-second throughput.

To keep the inference workers in sync with the rest of the training, we host a step counter endpoint which returns the smallest step with insufficient rollouts.
Inference workers poll this endpoint and generate rollouts for the step specified.
This design allows workers to dynamically join or leave the compute pool without interrupting the training process.

\subsubsection{Verifiers}

\primeframework \ uses the \genesys \ schema introduced in SYNTHETIC-1 \citep{2025synthetic1}, making it easy to implement new reward environments. In our initial experiments, we support symbolic verifiers for mathematics and unit test execution for python-based coding competition problems. For this, we adopt existing implementations from \citep{deepscaler2025} and \citep{cui2025processreinforcementimplicitrewards}.

Note that at this point, LLM-generated code is executed on the inference nodes, where we already apply sandboxing and code sanitization. This approach provides sufficient isolation for the simple algorithmic challenges currently used. For more complex coding tasks (e.g., those requiring filesystem access), further strengthening of isolation mechanisms will be necessary.

\subsection{\shardcast: Efficient Policy Weight Broadcasts}

\begin{figure*}
  \centering
  \includegraphics[width=0.85\linewidth]{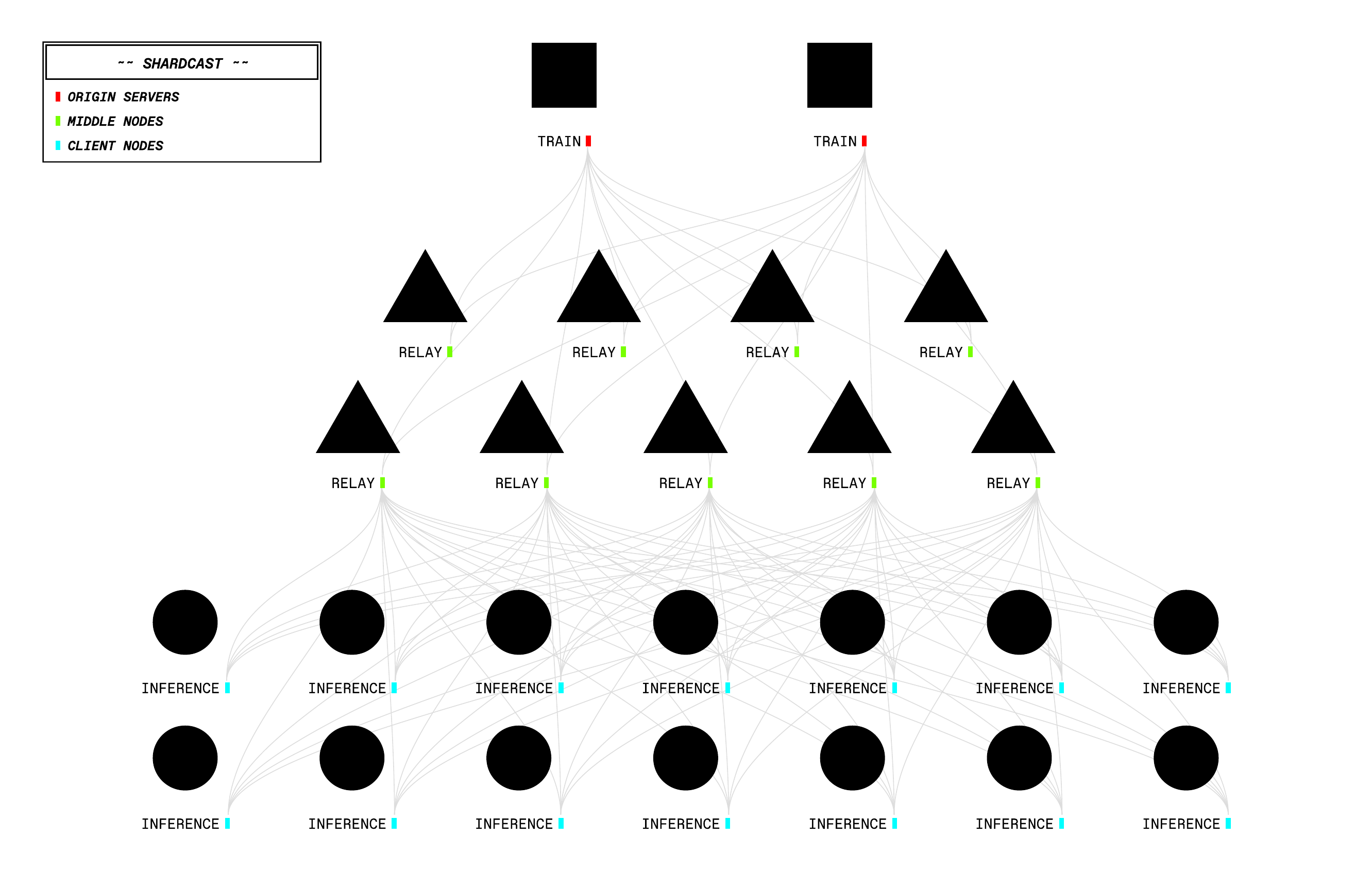}
  \caption{Overview of the Shardcast policy weight distribution network.}
  \label{fig:shardcast-topology}
\end{figure*}

One of the key challenges in asynchronous distributed reinforcement learning (RL) in a decentralized setting is ensuring that the most recent policy weights are quickly delivered to the inference workers.

For this purpose, we utilize a network of relay servers that distribute the checkpoints from the main training servers to the client inference workers, similar to a content delivery network (CDN).
To minimize latency, checkpoint files are sharded and streamed in a pipelined fashion, allowing inference workers to begin downloading shards before the full checkpoint is available on the relay servers.
The relay servers only keep the last five checkpoint versions to avoid running out of disk.
This limit is also functionally desirable, as rollouts generated with outdated checkpoints are typically rejected or discarded.

\subsubsection{Rate Limiting \& Firewall}

We use nginx \footnote{https://nginx.org/} as our HTTP server due to its robustness and widespread industry adoption.
To protect the relay servers from malicious inference workers who may make an excessive number of requests, we configure our nginx server with per IP rate limiting.
Additionally, we dynamically configure UFW firewall rules on the relay servers to accept traffic only from known, currently active inference nodes in the compute pool.
This reduces the surface area for attacks and enables us to quickly blacklist misbehaving nodes when detected.

\subsubsection{Maximizing Client Throughput \& Load Balancing}

If each client were to always select the fastest relay server, it would lead to contention and bandwidth thrashing.
To mitigate this, clients instead sample from the set of relay servers based on the expected throughput of requesting from each relay server.

Our implementation begins by having each client request a dummy file from all relay servers to initialize bandwidth and success rate estimates. Clients then select servers in proportion to:

$$
\text{expected throughput} \propto \text{success rate} \times \text{bandwidth}
$$

These estimates are continuously updated using an exponential moving average (EMA), which smooths transient fluctuations while remaining responsive to actual changes.
A healing factor is incorporated into the EMA to encourage periodic exploration of underutilized servers, ensuring the system adapts effectively to changing conditions.

Even in scenarios without contention, this probabilistic sampling strategy outperforms greedily choosing the currently fastest relay because it is able to utilize multiple connections to different relay servers, which will have a higher total bandwidth than any single connection to a relay.

\subsubsection{Assembled Model Weights Integrity Checks}
Inference nodes face the risk of being penalized if they submit rollouts generated using incorrect model weights.
To avoid this, it is essential that only verified, correct weights are used for inference.

To ensure model weight integrity, each inference worker computes the SHA-256 checksum of the assembled checkpoint after downloading and reconstructing it from the shards.
This checksum is then compared against the reference checksum produced by the training nodes, which is broadcasted to all relay servers along with the checkpoint metadata.

If the computed checksum does not match the expected value, the inference node discards the corrupted checkpoint and proceeds to attempt download of the next available checkpoint.
We avoid retrying the same checkpoint, as it is unlikely the re-download would complete before the checkpoint becomes stale or irrelevant.

While a mechanism for verifying the integrity of downloaded shards would enable trustless peer-to-peer (P2P) weight transfers, we chose not to deploy such a system for \intellect.
The primary reason is the added complexity and security risks associated with exposing inference workers to each another.
In a P2P setup, inference worker IP addresses would become visible to peers, requiring additional hardening to prevent malicious behavior or denial-of-service attacks within the pool.
Given these concerns, we opted to centralize weight distribution through trusted relay servers instead.

\subsection{\toploc: Enabling Trustless Inference}
\label{toploc_section}

\begin{figure*}
  \centering
  \includegraphics[width=\linewidth]{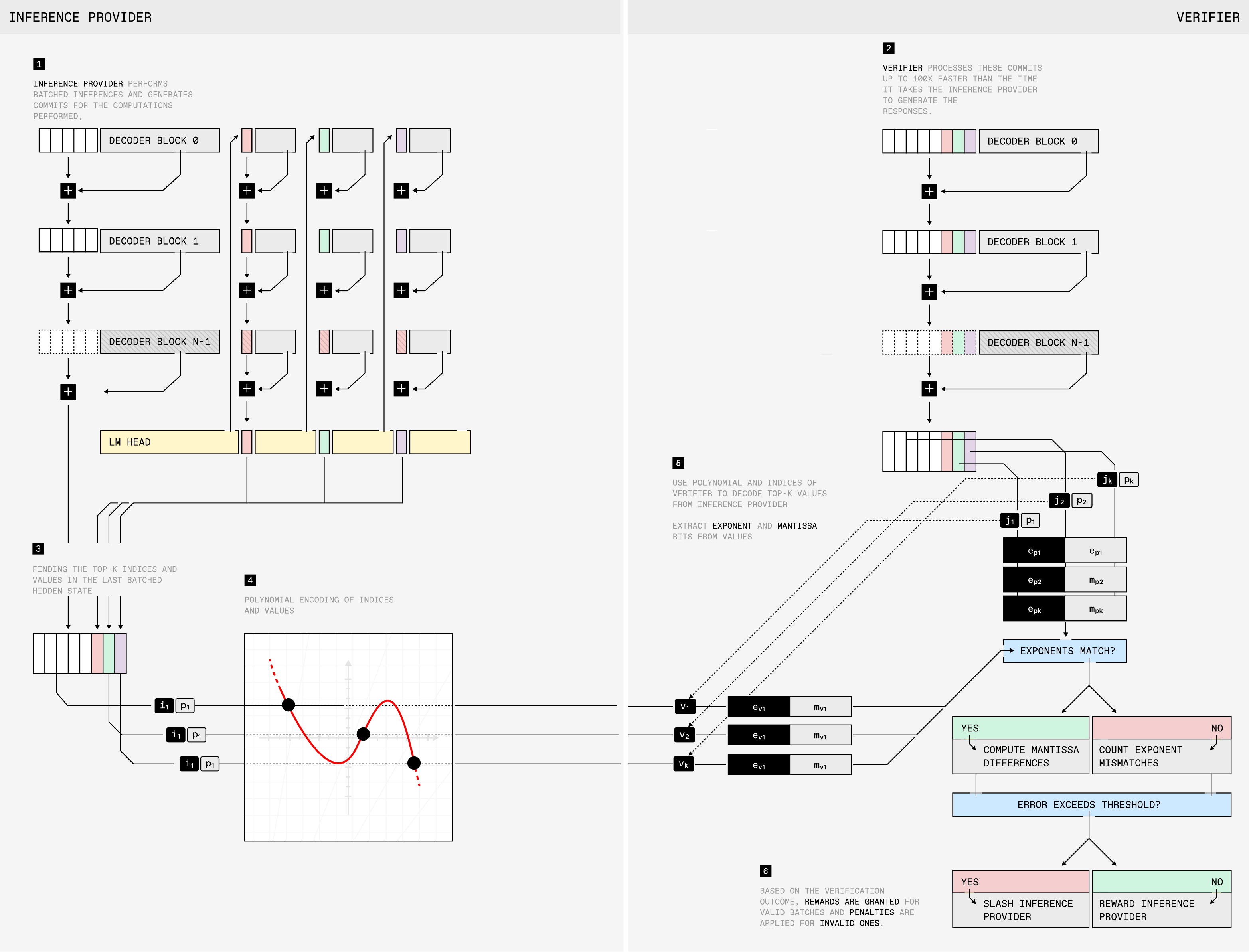}
  \caption{An illustration of \toploc. The Inference Provider performs batched inferences and generates commits for the computations performed, while the Verifier audits these commits up to $100\times$ faster than the time it takes the inference provider to generate the responses. Based on the verification outcome, rewards are granted for valid batches and penalties are applied for invalid ones. Further speedup can be obtained for the Verifier by not checking every batch but instead sampling randomly. Since the Inference Provider does not know which generations will be checked by the Verifier, they are incentivized to be honest on all generations to collect the reward and avoid receiving the penalty.}
  \label{fig:toploc}
\end{figure*}

Because we rely on trustless compute nodes for inference, there is no inherent guarantee that inference nodes perform the inference faithfully.
To ensure verifiable compliance, our validators use several checks: computation, sampling and data sanity check which we will describe in depth in the following sections.

\subsubsection{Computation checks}

\paragraph{Proof of correct model computation}
As shown in Figure \ref{fig:toploc}, to confirm that inference was performed using the correct model weights, each inference worker generates a \toploc\ proof \cite{toploc} for every generated sequence.
These proofs serve as cryptographic commitments to the final hidden states produced during decoding.
A trusted validator node subsequently reconstructs these activations using prefill and compares them to the submitted commitments to confirm consistency.
The proofs generated with \toploc \ are robust against GPU non-determinism, different tensor parallel configurations and are able to reliably detect when quantized or malicious versions of the model are used.

\subsubsection{Sampling checks}
\paragraph{Termination check}
There are two valid termination criteria for generated sequences: reaching the model’s maximum context length or producing an end-of-sequence (EOS) token.
Since longer sequences incur greater computational cost, inference providers may be incentivized to terminate sequences prematurely.
To guard against this, we check that either the sequence reaches the maximum model length or ends with an EOS token.
In the case that the sequence terminated on the generation of an EOS token, we make sure that the EOS token’s probability exceeds $0.1$ to prevent manipulation through unlikely EOS generations.

\paragraph{Token sampling check}
Proper sampling from logits should yield a distribution resembling an exponential with a mode at 1.
If a smaller model is used to generate tokens and only the larger model is used for prefill (to pass \toploc\ checks), the resulting distribution becomes bimodal, with modes near 1 and 0.
We inspect the logit distribution to detect such inconsistencies.

\subsubsection{Sanity checks}
\paragraph{Fixed data sampling}
\label{fixed_data_sampling}
Allowing inference workers to choose their own samples could lead to cherry-picking easy or previously completed examples.
To prevent this, each node deterministically selects samples based on a seed computed as:

\[
\text{seed} = \text{node address} \cdot \text{step} + \text{number of submissions for this step}
\]

We then verify that the correct samples were used by reproducing the sampling process from the seed.

\paragraph{Value bounds check}
All reported scalar values—such as rewards and advantages—must fall within predefined bounds to ensure they are reasonable and consistent with expected outcomes.

\paragraph{Parquet formatting check}
We also make sure that the parquet has the correct schema and is in a format that is loadable by our training dataloader.
This makes sure that we do not accept any files that would throw exceptions in the trainer.

\subsection{The Prime Intellect Protocol}
The Prime Intellect protocol coordinates permissionless nodes through a modular, decentralized orchestration layer.
It gives model trainers the ability to check the health of all nodes, view logs, and distribute new tasks analogous to a decentralized SLURM\footnote{Simple Linux for Resource Management: \url{https://slurm.schedmd.com/}}. The entire codebase is open source and available on GitHub\footnote{Prime Intellect Protocol: \url{https://github.com/primeIntellect-ai/protocol}}.

\subsubsection{System Architecture} 
\begin{figure*}
  \centering 
  \includegraphics[width=\linewidth]{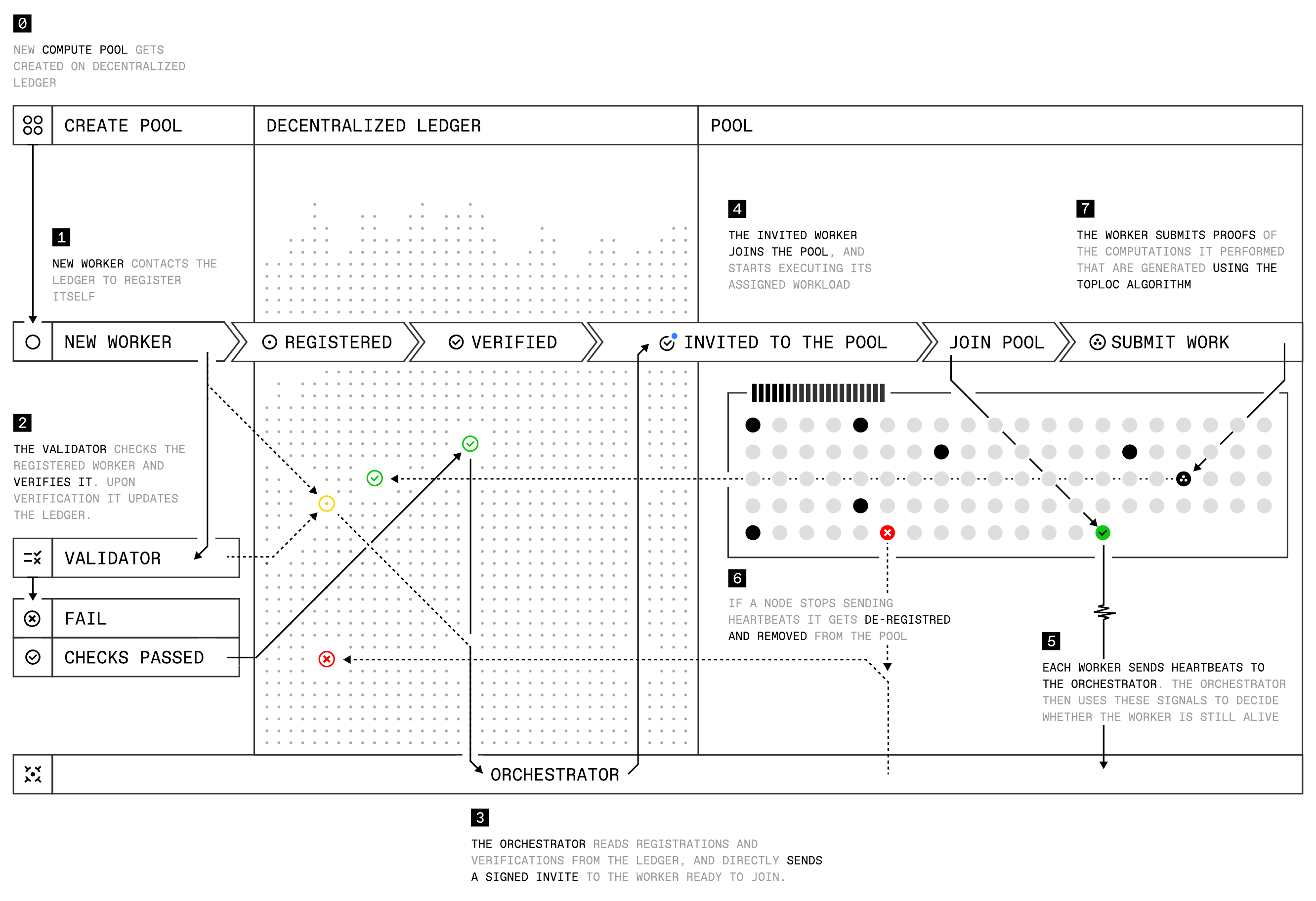}
  \caption{Overview of Protocol Testnet Infrastructure.}
  \label{fig:protocol-topology}
\end{figure*}

As shown in \autoref{fig:protocol-topology}, the system is composed of multiple components that are all implemented in Rust. All components, except for the worker nodes and decentralized ledger, are hosted in a Kubernetes cluster. All API endpoints are protected by Cloudflare.

\paragraph{Decentralized Ledger}
A decentralized ledger is used to store information about the current training run, ownership of the training run, as well as worker contributions. These workloads are organized into "compute pools" that all belong to a broader "compute domain." Each compute domain represents a specific category of AI task—such as pre-training, synthetic data generation, distributed reinforcement learning, and more.

The ledger maintains detailed information about each pool, including ownership details and worker contributions. Each contributor, as well as the compute pool owner, has a cryptographic address used for signing transactions and proving ownership, which secures API interactions and ensures proper attribution of compute resources.

\paragraph{Worker Software}
The worker software's core task is to communicate heartbeats and metrics to the central orchestrator and to configure and manage the local Docker environment for task execution. Additional features include exports of logs and restart capabilities for running containers, a Unix socket-based connection between the Docker container itself and the worker. The latter can be used to trigger actions on the worker software, such as file uploads from the Docker volumes. 

\paragraph{Discovery Service}
The discovery service is a simple API that allows nodes to upload worker metadata information. It stores this data in a Redis database and allows other authorized components, such as the orchestrator to retrieve information about nodes that have signed up. This ensures worker IPs are only visible to the orchestrator, reducing the risk of denial-of-service attacks.

\paragraph{Orchestrator}
The orchestrator's core tasks include the distribution of tasks and observing the lifecycle of the decentralized worker nodes based on their heartbeat. It gives the model trainer the ability to interact with the infrastructure via web API. It thereby exposes information about all nodes that are currently alive, an API to create and schedule new tasks, and insights into the current metrics and logs of each node. Additionally, since containers on each node might fail, each node's workload can be restarted.

\subsubsection{Operational Flows} 

\paragraph{Node Registration \& Discovery}  
The worker software is installed and started by compute contributors on their machines. It automatically detects the system components (GPU, available RAM and storage) and checks these for compatibility. Additional software and connection uplink checks are performed and inform the user in the logs about any misconfigurations or system issues that make the node incompatible with the training run.

Once the system hardware and software are confirmed, the node automatically uploads its metadata, including hardware information and IP, to a discovery service. In parallel, the node also sends a registration call to the decentralized ledger. After successful registration, the worker starts a webserver and waits for an invitation - this security measure ensures the worker doesn't need to know the orchestrator's endpoint in advance, protecting the orchestrator from potential denial-of-service attacks.

The orchestrator periodically checks the discovery service for newly created nodes and sends an invite to the worker's HTTP server to start contributing. This invite contains a cryptographic signature combining the node's address as well as the current compute pool's ID and domain. The invite is validated on the decentralized ledger and makes the worker an active compute contributor. After sending the invite, the orchestrator stores the node information in the local Redis storage and waits for incoming heartbeats. 

\paragraph{Node Health \& Heartbeats} 
Each node maintains a continuous heartbeat loop to maintain communication with the orchestrator. These heartbeats act as simple signals sent from the node back to the orchestrator, allowing it to track whether nodes are still active. The orchestrator stores these heartbeats in Redis with an expiration time, so it can automatically detect when a node stops responding. A separate status update loop regularly checks the health of each node by counting missed heartbeats. If too many are missed, the node is marked as dead, and its worker is removed from the decentralized ledger. If the node comes back online, it tries to re-register and update the discovery service so it can be invited back into the pool. 

\begin{figure*}
  \centering
  \includegraphics[width=\linewidth]{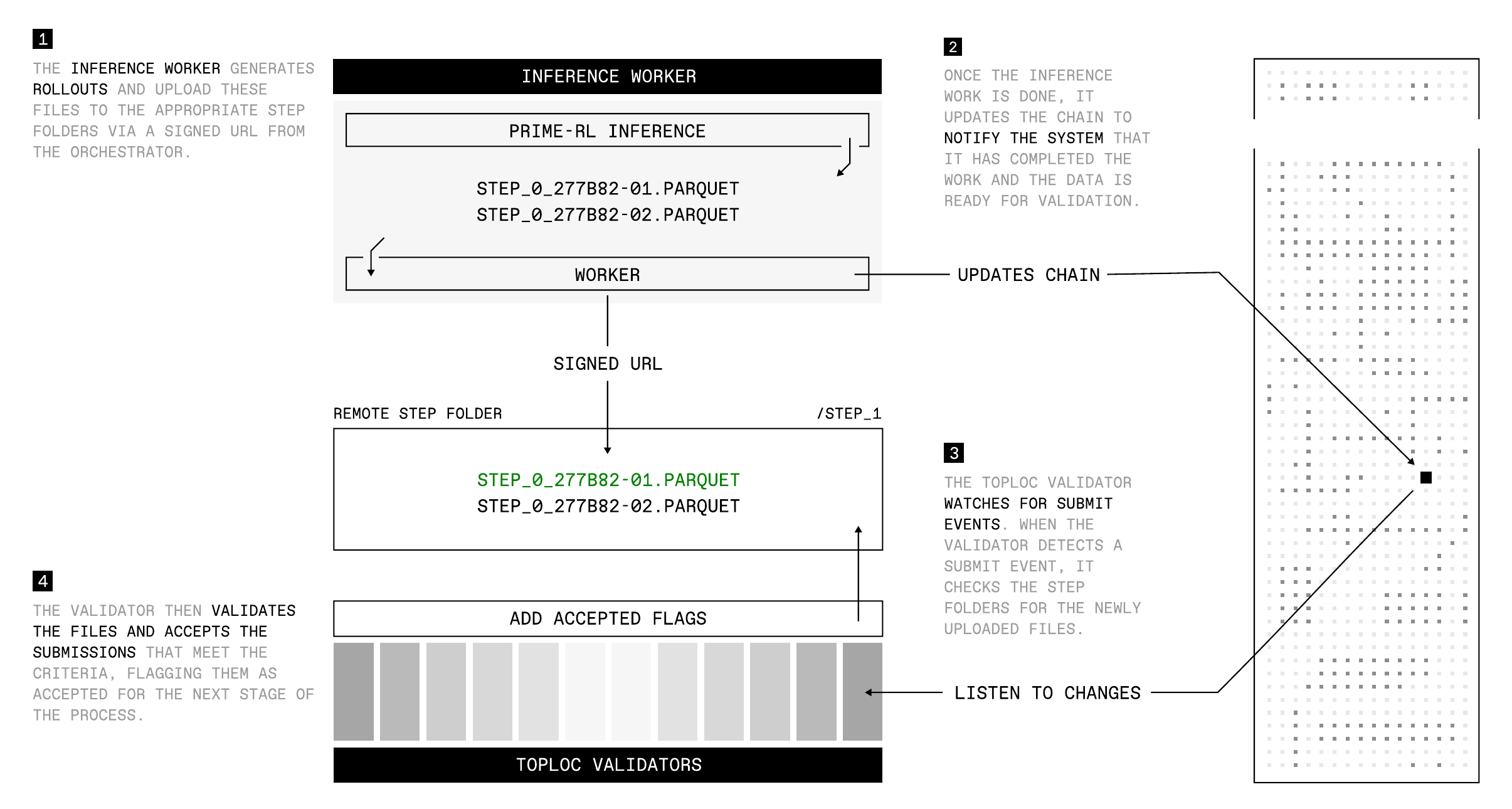}
  \caption{Overview of \toploc\ Validator Setup.}
  \label{fig:toploc_validator}
\end{figure*}

\paragraph{Task Scheduling \& Execution}
Tasks are created by the orchestrator through a POST API and scheduled asynchronously across all healthy nodes. Rather than pushing tasks, the orchestrator distributes them in response to heartbeat requests from the nodes, allowing for a reactive and fault-tolerant pull-based model. 
Once a task is received, the worker communicates with the Docker daemon to translate the task specification into a running container — this includes setting up volumes, managing the container lifecycle, and applying task-specific settings such as environment variables and custom start commands. One key insight during development was the introduction of a shared volume, used to store persistent data like model weights. Without this, restarting a task would trigger redundant downloads, slowing execution and increasing resource usage.

\paragraph{Inference Validation}
As shown in Figure \ref{fig:toploc_validator}, the inference workers will generate the rollouts and upload them to the remote folder using a signed URL.
An on-chain event is then generated, which triggers the validators to begin validation for the new file.
Based on the result of the checks described in Section \ref{toploc_section}, the file is either accepted or rejected.
The accepted files are then read by the training node data loaders.
Rejected files cause the node which created to them to be slashed and evicted from the compute pool.

\subsubsection{Design Trade-offs \& Limitations}
The orchestrator and discovery service are currently centralized, which simplifies coordination but creates potential single points of failure and limits horizontal scalability. This design also introduces trust assumptions that may not be suitable for more distributed or permissionless environments. To address this, we plan to move toward a fully peer-to-peer architecture using a distributed hash table (DHT), which would eliminate the need for central coordination and enable more resilient, decentralized node discovery.

Another key limitation lies in how the worker is deployed. At present, it only runs on bare-metal machines or virtual machines with direct access to the Docker daemon. This excludes environments like Kubernetes, where Docker access is abstracted or unavailable. To address this, we are developing a version of the worker that can run as a container itself, making it compatible with container orchestration platforms.

\section{Training Recipe}
\label{sec:training-recipe}

The goal of INTELLECT-2 is to train a model with reasoning capabilities, specifically in the domains of mathematics and coding. Additionally, we aim to enable control over the model's thinking budget by allowing users to specify the desired number of thinking tokens as part of the task prompt. As our base model, we use QwQ-32B \citep{qwq32b} and largely follow Deepseek-R1's \citep{deepseekai2025deepseekr1incentivizingreasoningcapability} approach of GRPO-based training with verifiable rewards.

In this section, we describe our RL recipe and the ablation experiments that led to it in detail, ranging from our training data and reward function implementations to modifications to the original GRPO objective to improve training stability.

\subsection{Training Data \& Rewards}

We train INTELLECT-2 using a dual objective: we incorporate both task rewards encouraging the model to improve its reasoning on mathematics and coding tasks, as well as length rewards in order to teach the model to adhere to a thinking budget provided in the prompt.

\subsubsection{Task Rewards}

Following existing state-of-the-art open reasoning models \cite{deepseekai2025deepseekr1incentivizingreasoningcapability, qwq32b} , we curate a training dataset consisting of mathematics and coding tasks that can be verified through symbolic verification / string matching and unit test execution.
To do so, we choose high quality problems from NuminaMath-1.5 \cite{numina_math_datasets} and Deepscaler \cite{Meng2023DeepScalerHA} for math problems and coding tasks previously curated for SYNTHETIC-1~\cite{synthetic1release2025}, which were also used in prior work such as DeepCoder \cite{deepcoder2025}. Our full dataset consists of 285k tasks, including 26k python-based algorithmic coding challenges and 259k mathematics problems. The dataset can be found on Huggingface \footnote{https://huggingface.co/datasets/PrimeIntellect/Intellect-2-RL-Dataset}. 

Both our mathematics and code reward functions implement binary rewards, with a reward of 1 being assigned for correct responses and 0 for incorrect responses.
While this is an obvious choice for mathematics tasks, we explicitly don't assign partial rewards for passing some, but not all unit tests of coding problems to discourage reward hacking (e.g. through memorizing public test cases).

\subsubsection{Length Rewards}

Beyond rewards for solving tasks correctly, we incorporate length rewards to enable users to specify the thinking budget of \intellect\ as part of the task prompt; hereby, we largely follow the methodology of L1 \citep{aggarwal2025l1controllinglongreasoning}.

Concretely, for every problem in a training batch, we sample a target length \(l_{\mathrm{target}}\) and include it in our prompt via the template "Think for \(l_{\mathrm{target}}\) tokens before giving a response." - subsequently, a length penalty representing the difference between the actual response length and the target length, multiplied by a weighting factor \(\alpha\), is combined with the task reward. Letting \(y\) denote our model output for a given prompt, \(l_y\) its length in tokens, and \(r_{\mathrm{task}}\) the task reward function, the total reward can be computed as:

\[r_{\mathrm{total}}(y, l_{\mathrm{target}}) = r_{\mathrm{task}}(y) - \alpha * | \ l_{\mathrm{target}} - l_y \ |\]

Different from the setting of L1, where \(l_{\mathrm{target}}\)  is sampled uniformly from a continuous range, we sample from a small discrete set of target lengths (e.g., 2000, 4000, 6000 tokens) to simplify the objective and make it easier for our model to learn. To validate this approach, we reproduced L1 using target lengths of 500, 1000, 2000 and 3000 with a maximum sequence length of 4000.

\begin{figure*}
  \centering
  \includegraphics[width=\linewidth]{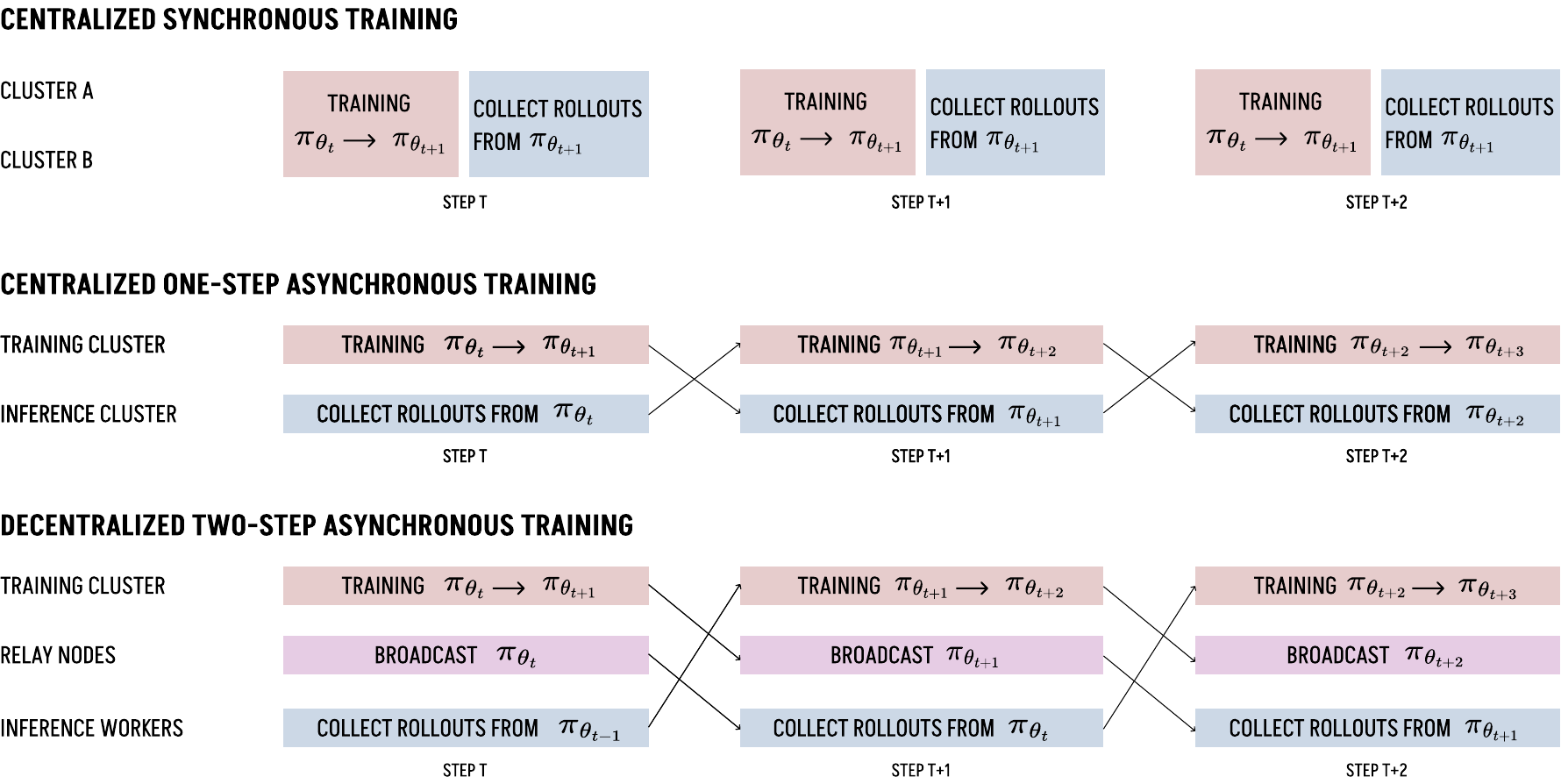}
  \caption{A comparison of synchronous, centralized one-step asynchronous and decentralized two-step asynchronous reinforcement learning: \textbf{Synchronous RL} leverages the same compute resources for training and inference and sequentially switches between performing only inference and training. Therefore, training is fully on-policy. \textbf{Centralized One-Step Asynchronous RL} has dedicated compute resources for training and inference and performs training and inference at the same time. Therefore, rollouts are collected from the policy of the last RL step, making training off-policy by one step. \textbf{Decentralized Two-Step Asynchronous RL} works similarly to centralized asynchronous RL, but inference workers don't have access to the up-to-date policy weights immediately after a training step due to the time-consuming weight broadcast. Therefore, rollouts are collected from policy weights from two or more RL steps prior.}
  \label{fig:async-rl-scheduling}
\end{figure*}

\subsection{Asynchronous Reinforcement Learning}

 As discussed in Section \ref{sec:infra}, we use asynchronous reinforcement learning to use both dedicated inference and training nodes to minimize GPU idle time. This approach has proven to be effective in prior work \citep{huang2023cleanbareproducibleefficientdistributed, noukhovitch2025asynchronousrlhffasterefficient} and has also been adopted for large LLM training runs such as Tülu 3 \citep{lambert2025tulu3} and Llama 4 \citep{llama4}.

In a centralized asynchronous RL training setup with fast connection speeds, the same policy weights that are updated on the dedicated training nodes are simultaneously used to obtain rollouts for training during the next RL step. In a decentralized setup, the updated policy weights are not available immediately to the inference workers, as the weight broadcast costs time, which is why we perform rollouts using weights not from the previous step, but from two or more steps prior, depending on the duration of the weight broadcast. A graphical overview of these differences along with a comparison to synchronous RL can be found in \autoref{fig:async-rl-scheduling}.

\begin{figure*}
  \centering
  \includegraphics[width=0.9\linewidth]{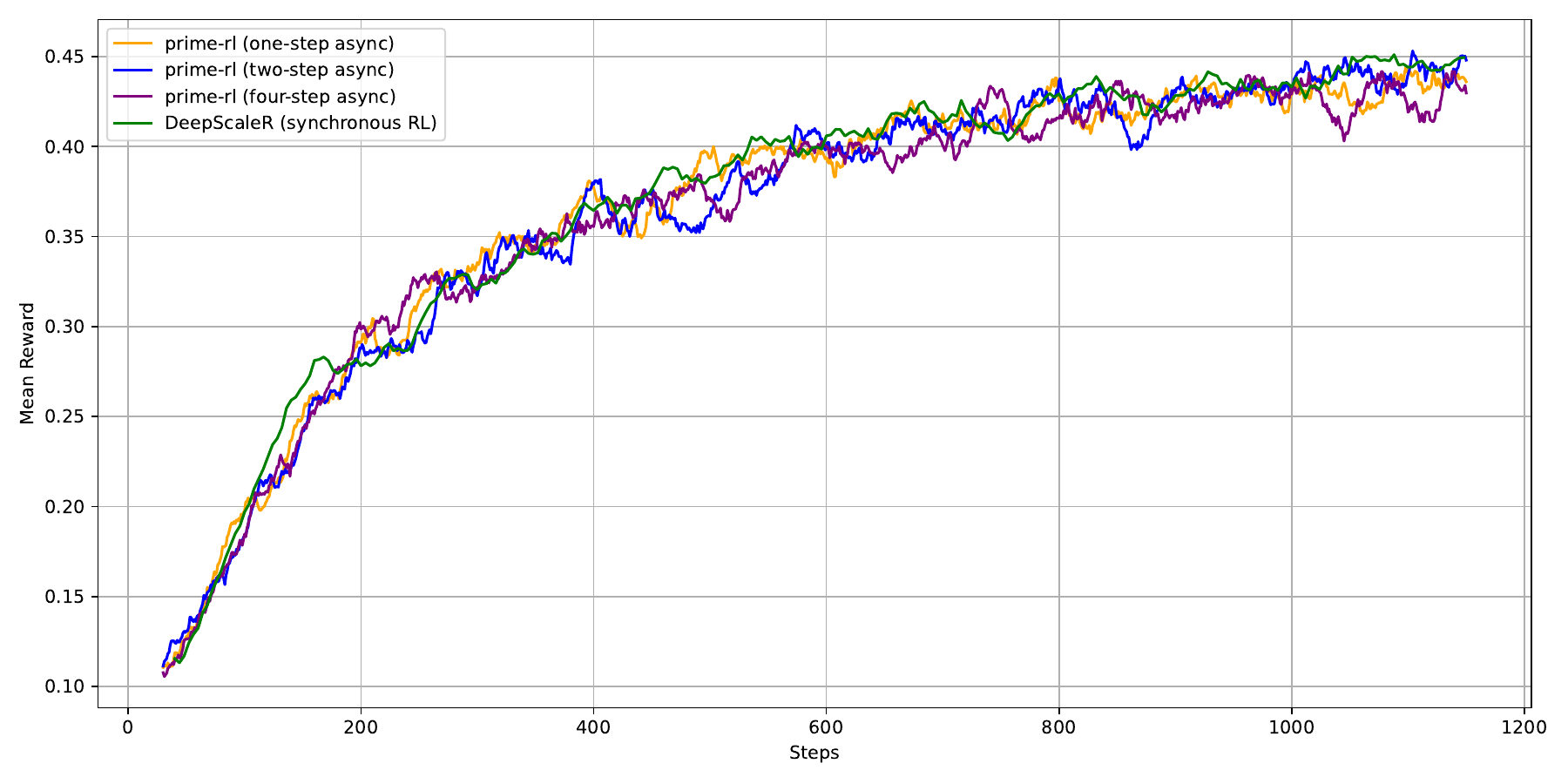}
  \caption{Comparison of synchronous DeepScaler \citep{Meng2023DeepScalerHA} training vs asynchronous \primeframework\ under varying asynchrony levels. Even with increased delay (up to four steps), \primeframework\ matches the performance of synchronous baselines.}
  \label{fig:asynchrony}
\end{figure*}

Prior to starting our \intellect\ training run, we ran ablation experiments to validate that asynchronous RL training does not hurt the performance of our model.
To do so, we replicated the results of Deepscaler's synchronous RL training run of DeepSeek-R1-Distill-Qwen-1.5B using a context length of $2048$ tokens and compared it with asynchronous RL based on \primeframework with varying levels of asynchrony.
The results of these runs can be found in Figure \ref{fig:asynchrony}.

As seen in the graph, even with asynchrony levels of up to four, our model's reward trajectory matches the trajectory of the synchronous baseline, indicating that training on slightly off-policy data does not hurt the performance of RL training.

\subsection{Offline \& Online Data Filtering}

During our ablation experiments, we found that filtering our dataset for difficulty had a significant impact on training performance. We employ both offline filtering before starting our training run and online filtering to selectively choose training samples from our rollouts.

\subsubsection{Offline Data Filtering}

\begin{figure*}
    \centering
    \begin{subfigure}{0.48\textwidth}
        \includegraphics[width=\textwidth]{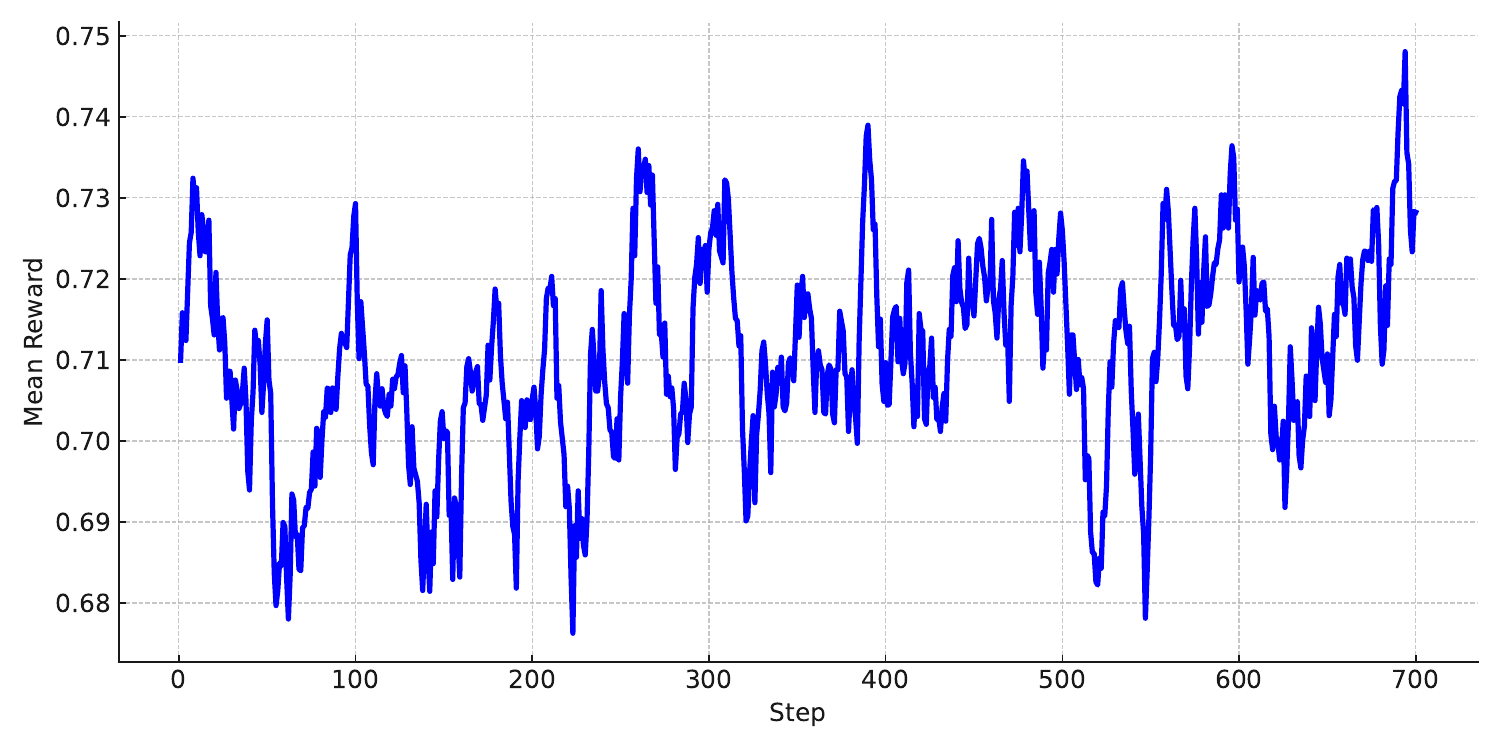}
        \caption{7B GRPO Training without Offline Filtering}
        \label{fig:deepscaler}
    \end{subfigure}
    \hfill
    \begin{subfigure}{0.48\textwidth}
        \includegraphics[width=\textwidth]{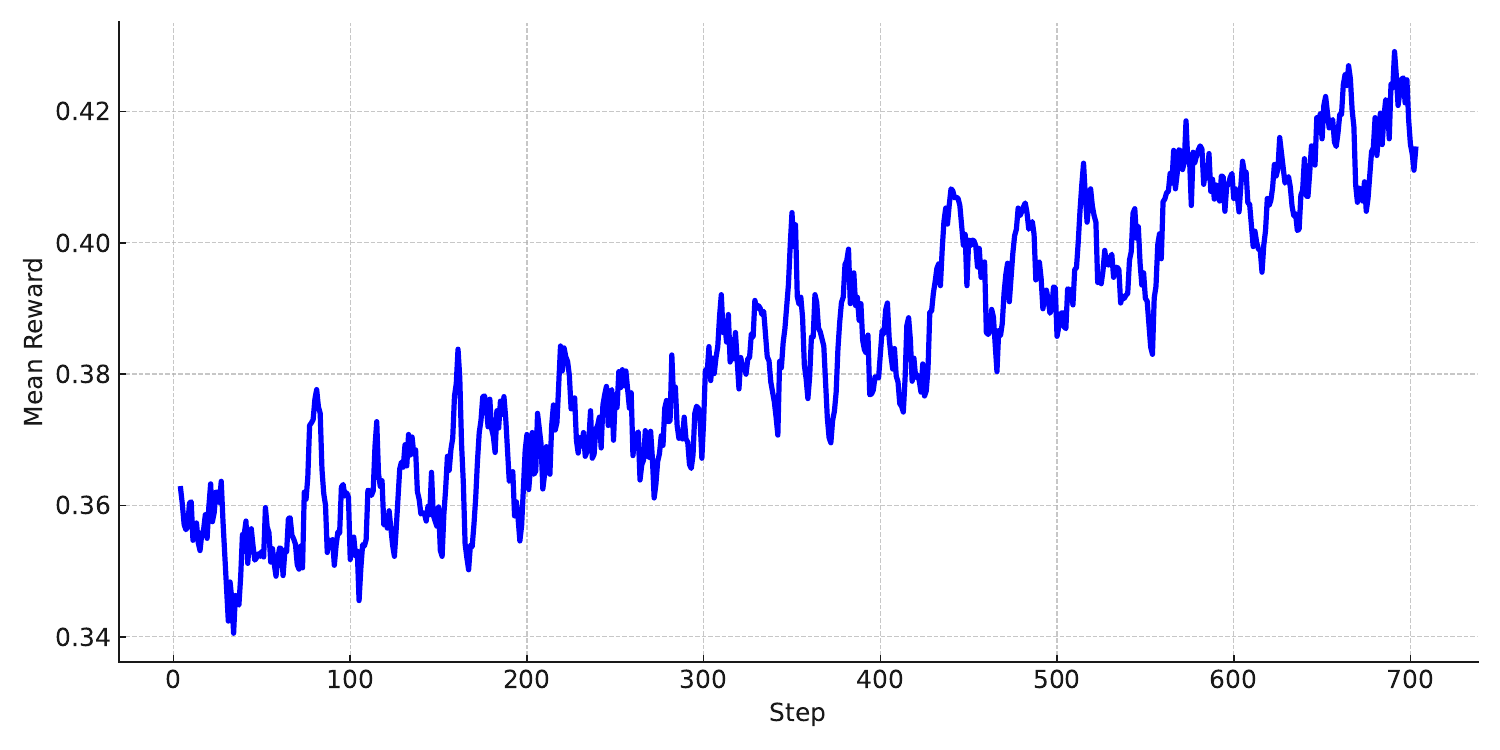}
        \caption{7B GRPO Training with Offline Filtering}
        \label{fig:deepscaler_filtering}
    \end{subfigure}
    \caption{Reward trajectories when training DeepSeek-R1-Distill-Qwen-7B using GRPO on the Deepscaler Math dataset with and without online filtering. \ref{fig:deepscaler} shows the reward trajectory when training without filtering, leading to stagnant rewards. \ref{fig:deepscaler_filtering} shows the results of training on a filtered dataset only containing samples on which the base model had a pass@8 value between 1 and 4.}
    \label{fig:difficulty_filtering}
\end{figure*}

When trying to train DeepSeek-R1-Distill-Qwen-7B using the Deepscaler mathematics dataset \citep{deepscaler2025}, we found that it was highly important to filter out problems that were too easy or too difficult from our training set. As shown in \autoref{fig:difficulty_filtering}, when training on the original dataset, our rewards barely improved. After filtering out problems in which the base model's pass@8 rate was above 50\%, and below 12.5\%, rewards improved significantly. As a result, we also used DeepSeek-R1-Distill-Qwen-7B to prefilter our training dataset for \intellect.

\subsubsection{Online Data Filtering}

Training algorithms such as GRPO \citep{shao2024deepseekmathpushinglimitsmathematical} and RLOO \citep{kool2019buy} rely on group-based relative rewards to compute advantages. If all completions for a single problem receive the same rewards (for binary rewards either 0 or 1), this means that the advantages for all of these samples are zero and no training signal is given beyond auxiliary losses such as a KL or entropy loss. To mitigate this, we employ online filtering and continue sampling responses from our inference workers until we have a full batch of samples with non-zero advantages before performing a training step. Conveniently, this increases the amount of inference that has to be performed per training step, allowing us to onboard and leverage a higher amount of decentralized inference nodes.

\subsection{Two-Sided GRPO Clipping for Increased Training Stability}

During training, we faced loss and resulting gradient norm spikes that caused instabilities leading to model collapse, particularly as our models got larger. Upon inspection, we found that a major cause of instabilities was one-sided token probability ratio clipping employed in GRPO and PPO-like \citep{schulman2017proximalpolicyoptimizationalgorithms} training objectives.

Recall the original GRPO objective: for each prompt or question \(q\), GRPO samples a group of outputs \(\{o_1, o_2, ...,o_G\}\) from the old policy \(\pi_{\theta_{\mathrm{old}}}\) and computes advantages based on their relative rewards based in this group. The optimization objective without the KL penalty is given by:

{\small
\[
\mathcal{J}_{\text{GRPO}}(\theta) = \mathbb{E}_{q \sim P(Q), \;\{o_i\}_{i=1}^{G} \sim \pi_{\theta_{\text{old}}}(O|q)}
\]
\[
\frac{1}{G} \sum_{i=1}^{G} \frac{1}{|o_i|} \sum_{t=1}^{|o_i|}
\left[
\min\left(
 \frac{\pi_\theta(o_{i,t} | q, o_{i,<t})}{\pi_{\theta_{\text{old}}}(o_{i,t} | q, o_{i,<t})}\hat{A}_{i,t},
\ \text{clip}\left(
\frac{\pi_\theta(o_{i,t} | q, o_{i,<t})}{\pi_{\theta_{\text{old}}}(o_{i,t} | q, o_{i,<t})},
1 - \varepsilon,\ 1 + \varepsilon
\right) \hat{A}_{i,t}
\right)
\right]
\]
}

where \(\hat{A}_{i,t}\) denotes the advantage within each sampled group and \(\varepsilon\) is a parameter used to clip the token probability ratio to avoid excessively large loss values and gradient updates. Note that in the case of negative advantage values, this clipping will not be applied due to the \(\mathrm{min}\) operation, as large updates that move the policy away from bad rollouts are encouraged \citep{schulman2017proximalpolicyoptimizationalgorithms}. However, this can cause huge loss values and gradient updates as a result in case \(\frac{\pi_\theta(o_{i,t} | q, o_{i,<t})}{\pi_{\theta_{\text{old}}}(o_{i,t} | q, o_{i,<t})}\) takes on large values.

To mitigate this, we introduce an additional hyperparameter \(\delta\) that adds an upper bound to the token probability ratio in the case of negative advantages:

{\small
\[
\mathcal{J}_{\text{GRPO}}(\theta) = \mathbb{E}_{q \sim P(Q), \;\{o_i\}_{i=1}^{G} \sim \pi_{\theta_{\text{old}}}(O|q)}
\]
\[
\frac{1}{G} \sum_{i=1}^{G} \frac{1}{|o_i|} \sum_{t=1}^{|o_i|}
\left[
\min\left(
 \color{red}\min\left(\color{black}\frac{\pi_\theta(o_{i,t} | q, o_{i,<t})}{\pi_{\theta_{\text{old}}}(o_{i,t} | q, o_{i,<t})}, \color{red}\delta\right)\color{black}\hat{A}_{i,t},
\ \text{clip}\left(
\frac{\pi_\theta(o_{i,t} | q, o_{i,<t})}{\pi_{\theta_{\text{old}}}(o_{i,t} | q, o_{i,<t})},
1 - \varepsilon,\ 1 + \varepsilon
\right) \hat{A}_{i,t}
\right)
\right]
\]
}

The value \(\delta\) should be higher than \(1+\varepsilon\) to still enable large updates that move away from bad rollouts, but avoid huge token probability ratios of a hundred or much higher. With this change, training stabilized significantly, as it has also been reported in concurrent work \citep{minimax2025minimax01scalingfoundationmodels}.

\subsection{Mitigating Training Instability at Scale}

While the two-sided GRPO clipping mechanism described above significantly reduced large loss and gradient spikes, we observed additional types of training instabilities when using larger models. These instabilities share similarities with those encountered in large-scale pretraining~\citep{wortsman2023smallscaleproxieslargescaletransformer, chowdhery2022palmscalinglanguagemodeling, molybog2023theoryadaminstabilitylargescale, cohen2024adaptivegradientmethodsedge}.

\begin{figure*}
    \centering
    \begin{subfigure}[b]{0.495\textwidth}
        \includegraphics[width=\textwidth]{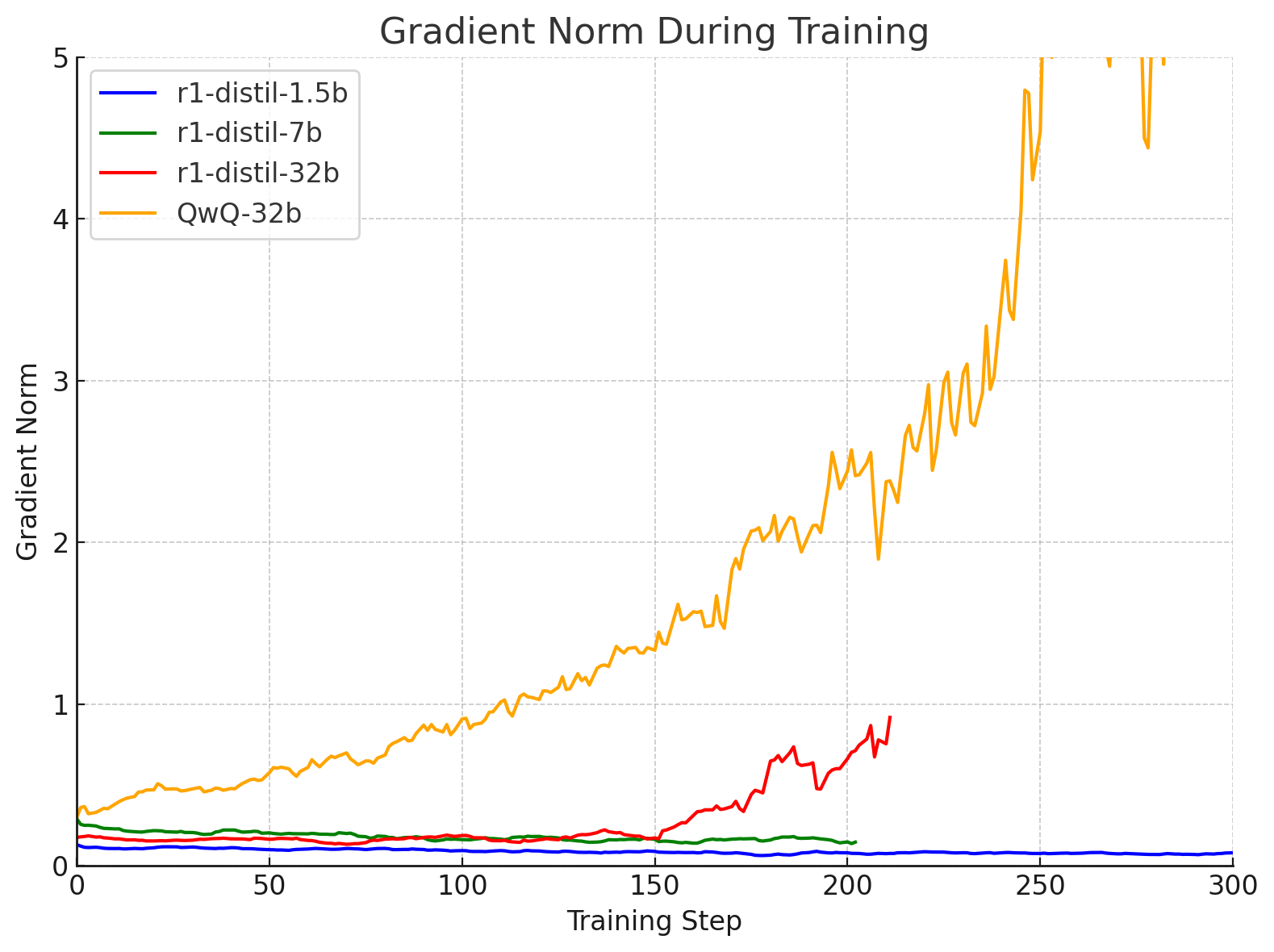}
        \caption{Gradient norms during training across training steps}
        \label{fig:grad_norms}
    \end{subfigure}
    \hfill
    \begin{subfigure}[b]{0.495\textwidth}
        \includegraphics[width=\textwidth]{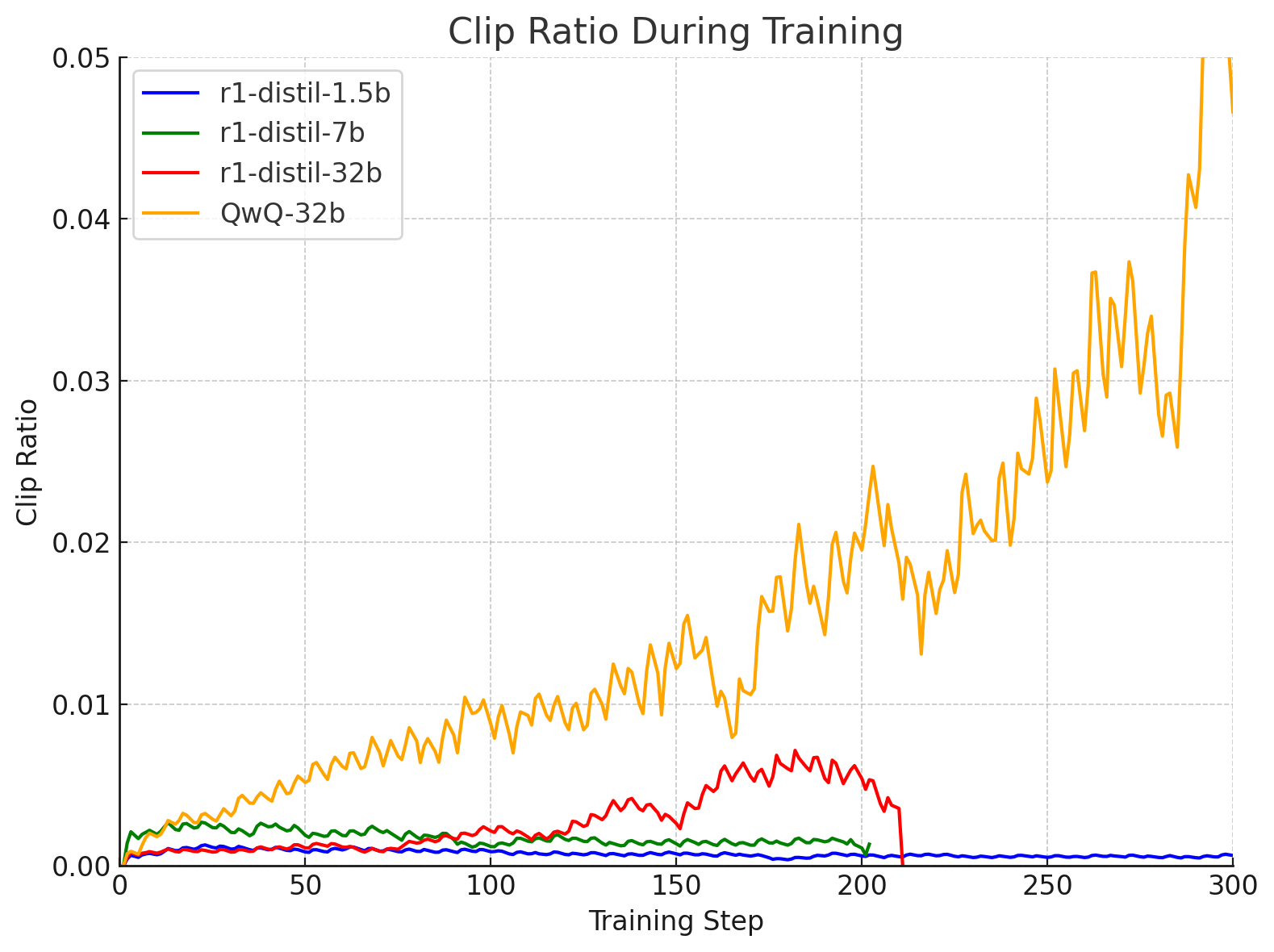}
        \caption{Token probability clipping ratio across training steps}
        \label{fig:clip_ratios}
    \end{subfigure}
    \caption{
        Escalating gradient norms (\autoref{fig:grad_norms}) and clipping ratios (\autoref{fig:clip_ratios}) across model scales, trained on the MATH dataset \citep{hendrycksmath2021}. Smaller models remain stable, while both 32B models show rising instability, with QwQ-32B diverging earlier than R1-Distill-32b.
    }
    \label{fig:grad_and_clip}
\end{figure*}

\paragraph{Escalating Gradient Norms}
As training progressed, we observed a gradual but persistent increase in gradient norms, even in the absence of immediate spikes. This phenomenon appears to be correlated with model size as shown in \ref{fig:grad_norms}, becoming more pronounced in our larger architectures. Similar to findings in \citep{wortsman2023smallscaleproxieslargescaletransformer}, we found that gradient norm growth often precedes more severe instability events, serving as an early warning signal for potential training collapse.

We found that employing aggressive gradient clipping (with thresholds as low as 0.05-0.1) effectively mitigates stability issues without significantly impeding convergence, providing a favorable trade-off between stability and training efficiency. While this approach does not completely eliminate instability issues, it substantially delays the growing gradient phase and postpones potential stability crashes, extending the viable training period for our models. Aggressive gradient clipping has also been applied successfully in concurrent work training QwQ-32B \citep{multi-turn-kernels}.

\paragraph{Token Probability Clip Ratio Escalation.} In addition to gradient norm and entropy instabilities, we observed a steady increase in the token probability clip ratio during training, as shown in \autoref{fig:clip_ratios}. This increase directly correlates with the rising gradient norm, as the clip ratio effectively tracks the difference in logits between consecutive optimizer steps.

\begin{figure*}
  \centering
  \includegraphics[width=0.8\linewidth]{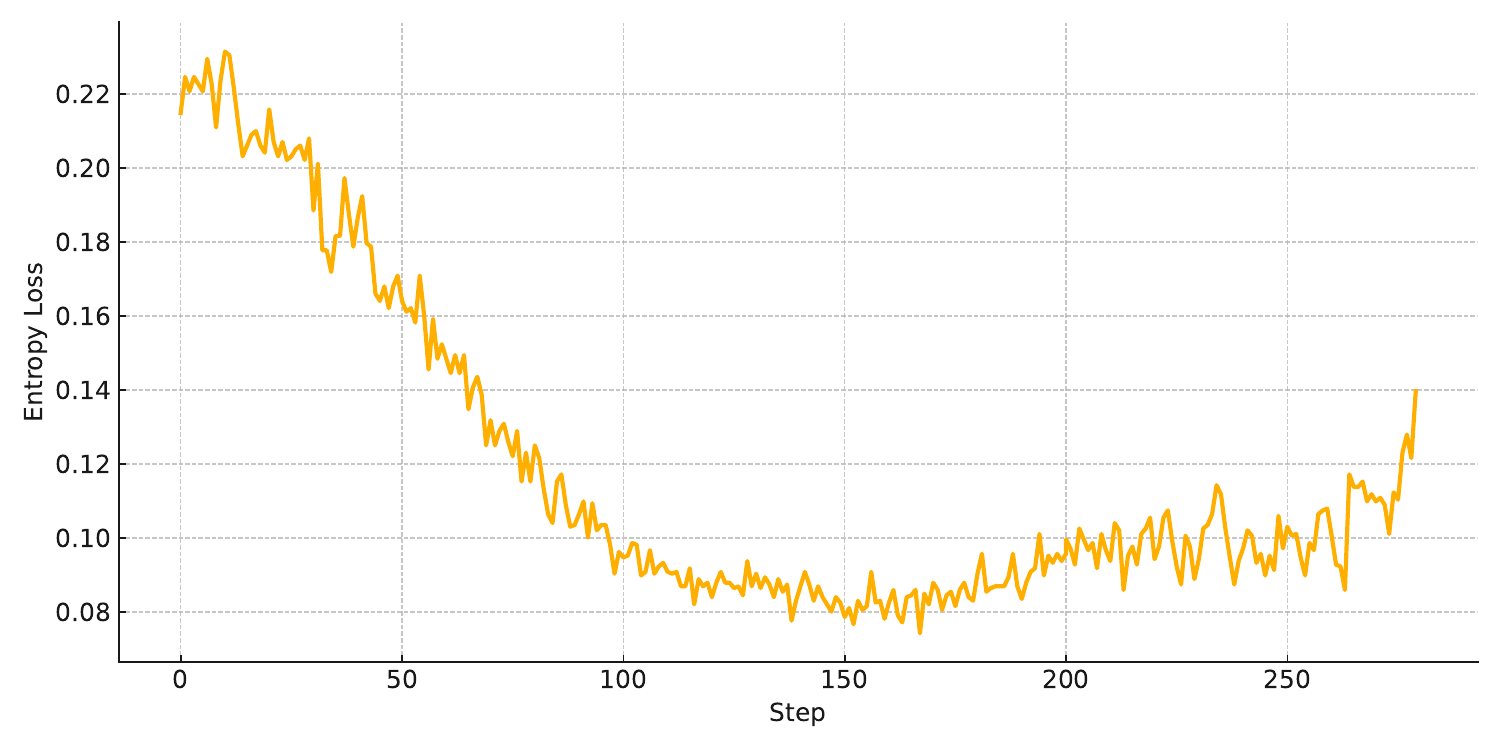}
  \caption{As training progresses, our policy's entropy loss initially decreases but later starts increasing past $\approx 150$ steps. Soon after seeing entropy increases, we observed our model collapsing across all of our ablation runs.}
  \label{fig:entropy-loss}
\end{figure*}

\paragraph{Entropy Loss pattern.} During training, we identified a distinctive pattern for the entropy loss, as shown in Figure \ref{fig:entropy-loss}. After decreasing initially, the entropy loss begins to trend upward again. This entropy resurgence typically precedes catastrophic training failures causing a full collapse of the model. Increasing the weighting factor of the KL penalty was able to delay this collapse but also caused slower learning, and hence wasn't an effective mitigation strategy.

\paragraph{QwQ is less stable to train than DeepSeek-R1-Distill-Qwen-32B.}
We noticed that our training on top of QwQ exhibited worse stability compared to DeepSeek-R1-Distill-Qwen-32B, despite both being based on the same pre-trained model (Qwen 2.5). We hypothesize that this difference stems from QwQ having already undergone a phase of reinforcement learning with verifiable rewards. This prior RL training appears to make the model more susceptible to subsequent optimization instabilities, suggesting that models may become progressively more difficult to fine-tune stably after multiple rounds of reward optimization.

\begin{figure*}
    \centering
    \begin{subfigure}[b]{0.48\textwidth}
        \includegraphics[width=\textwidth]{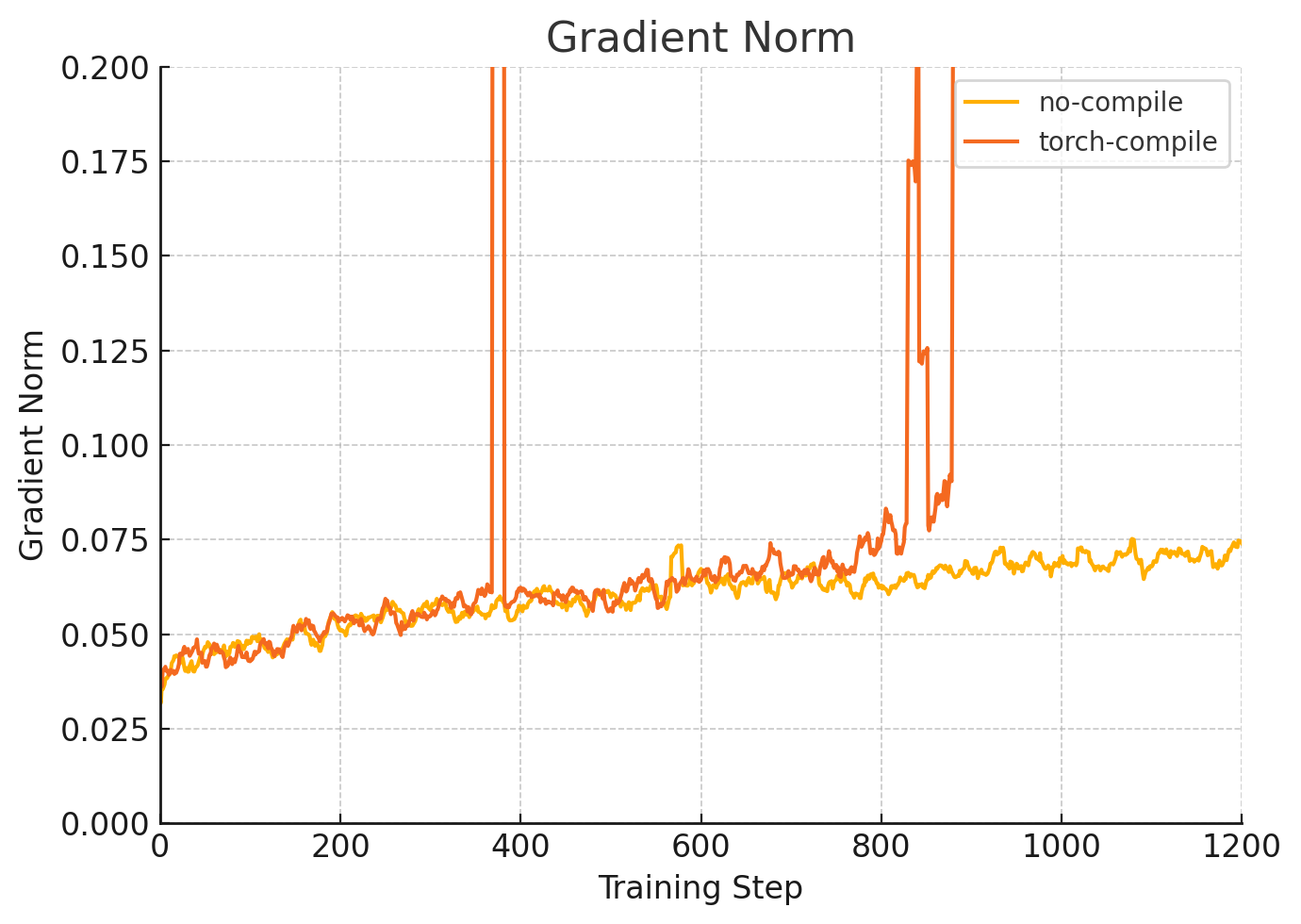}
        \caption{Gradient Norm}
        \label{fig:compile_grad_norm}
    \end{subfigure}
    \hfill
    \begin{subfigure}[b]{0.48\textwidth}
        \includegraphics[width=\textwidth]{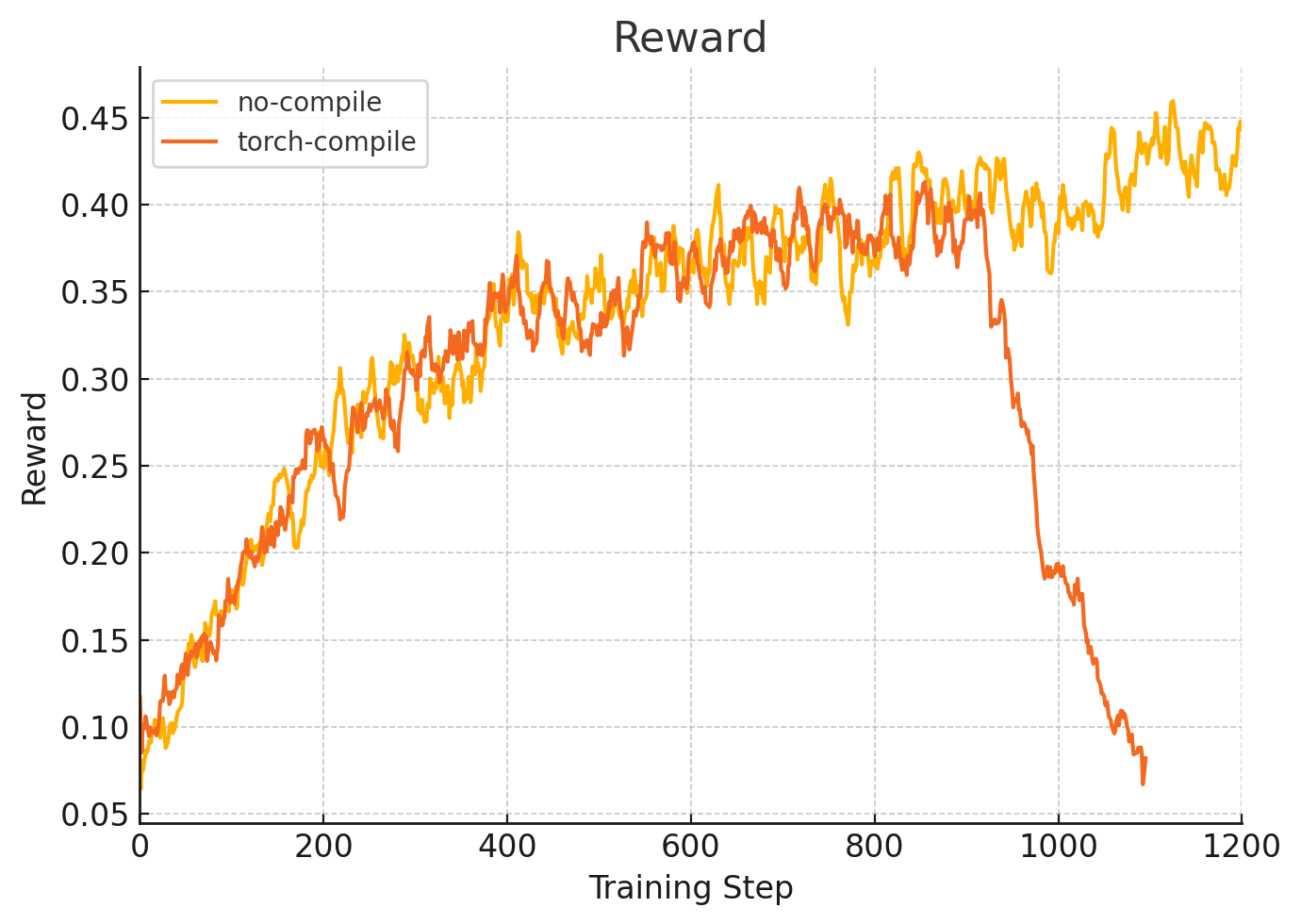}
        \caption{Reward}
        \label{fig:compile_rewards}
    \end{subfigure}
    \caption{
        Training dynamics with and without \texttt{torch.compile} when training DeepSeek-R1-Distill-Qwen-1.5B on the Deepscaler mathematics dataset. The \textit{torch-compile} setup leads to early instability and reward collapse, whereas the \textit{no-compile} baseline remains stable across 1200 steps.
    }
    \label{fig:torch_compile_grad_reward}
\end{figure*}

\paragraph{Instabilities caused by torch.compile.}
We observed that using \texttt{torch.compile} led to a catastrophic collapse during the later stages of our training, regardless of our model size (see \autoref{fig:torch_compile_grad_reward}). Although the issue likely stemmed from a single faulty kernel generated by \texttt{torch.compile}, we ultimately decided to disable it across the entire codebase for the run to ensure training stability. This decision came with a trade-off of slightly increased memory usage.

\section{Experiments}
\label{sec:experiment}

Over the duration of two weeks, we ran multiple training runs using our setup consisting of a trusted training cluster and validator nodes and trustless, community-contributed heterogenous inference workers. In this section, we report the experiments we ran along with corresponding results.

\subsection{Experimental Setup}

Using QwQ-32B as our base model, we trained using GRPO with modifications described in \autoref{sec:training-recipe} and used clipping thresholds \(\varepsilon = 0.2\), \(\delta = 4\) and an entropy loss coefficient of 1e-4. We set the KL divergence loss coefficient to 0.001, set \(\alpha\) to 0.0003 to balance task and length rewards, and apply gradient norm clipping at 0.1. Additionally, we implement a token-level policy loss calculation rather than a sample-level loss calculation as proposed in DAPO \citep{yu2025dapoopensourcellmreinforcement} and Dr. GRPO \citep{liu2025understandingr1zeroliketrainingcritical}. The training used a learning rate of 3e-7 with 25 warmup steps; during every rollout step, we generated 4096 samples consisting of 16 responses to 256 prompts, and performed 8 optimizer steps using a batch size of 512. We used two-step asynchrony during training to enhance throughput while ensuring that we did not go too far off-policy. 

We used the Huggingface implementation of Qwen with Flash Attention 2. Instead of obtaining token log probabilities from our inference workers, we computed them seperately using our policy weights on the training cluster, as vLLM log probabilities turned out to not be numerically stable. The model was sharded using FSDP2 with activation recomputation enabled. We configured a maximum sequence length of 32K to accommodate longer inputs required by our application.

\paragraph{Sequence Packing}
To maximize computational efficiency with our 32K sequence length, we implemented sequence packing to address the significant variance in sample lengths. This approach prevents wasting compute on padding tokens, which would be particularly inefficient given our distribution of sequence lengths during inference rollout. Unlike pretraining scenarios, where arbitrarily cutting samples is acceptable due to the local nature of the next token prediction loss function, reinforcement learning requires preserving complete samples since RL fundamentally learns at the sample level rather than locally. While RL fundamentally requires preserving complete samples, GRPO's token-level loss formulation allowed us to implement cross-sample packing by adapting the attention mask and collating samples into the sequence dimension. This optimization proved essential for scaling beyond 20K+ sequence lengths and significantly reduced our training time while maintaining the integrity of the cross entropy calculations across packed sequences.

\subsection{Results}

We report results from two main experiments: \shortexperiment, an experimental run with target lengths  \(\{1000, 2000, 3000, 4000\}\) to train an efficient reasoning model, and, \longexperiment, our main run with longer target lengths of \(\{2000, 4000, 6000, 8000, 10000\}\).

\paragraph{Compute Utilization}

During the two main experiments, we successfully overlapped communication with computation through asynchronous reinforcement learning.

In both experimental settings, the \shardcast\ broadcast to all nodes averaged 14 minutes, corresponding to a bandwidth throughput of approximately 590 Mb/s (62 GB of weights transmitted over 14 minutes).
Nodes with superior connectivity to the \shardcast\ relay servers received checkpoints earlier, allowing them to begin data generation ahead of others.
Furthermore, nodes with more computational resources, such as full H100 nodes, generated batches more quickly, resulting in earlier validation by the \toploc\ validators.

In the \shortexperiment\ setup, the first data file was submitted approximately 10 minutes after the broadcast completed.
Thanks to the prefill verification mechanism in \toploc\ and the random sampling strategy that verifies only subsets of submitted data, the inference verification was highly efficient—typically completing within 1 minute.
Consequently, sufficient verified samples to form a batch were available roughly 22 minutes after the broadcast in the \shortexperiment\ scenario.
In contrast, the \longexperiment\ scenario required approximately 29 minutes to accumulate enough verified samples to form a batch.

The ratio of training to inference FLOPs in both experiments averaged \(4.5\times\), with significantly more compute spent on the decentralized inference workers than on the training side.

In \shortexperiment, training nodes took approximately 22 minutes to execute a full rollout step.
This duration resulted in minimal idling of training GPUs, as the data generated by inference nodes for the next step was already available.
Conversely, in \longexperiment, training nodes completed their rollout steps in about 21 minutes. The asynchronous setup effectively synchronized with the broadcast, inference generation, and verification phases, ensuring nearly perfect computational overlap and minimizing GPU idling time.

These results highlight the inherent advantage of decentralized reinforcement learning, especially when scaling inference computations—such as generating longer reasoning chains—to achieve an optimal inference-to-training compute ratio. Additional analysis on this scaling benefit is discussed in \autoref{sec:discussion-decentralized-ai}.

\begin{figure*}
  \centering
  \begin{subfigure}{0.45\textwidth}
    \includegraphics[width=\linewidth]{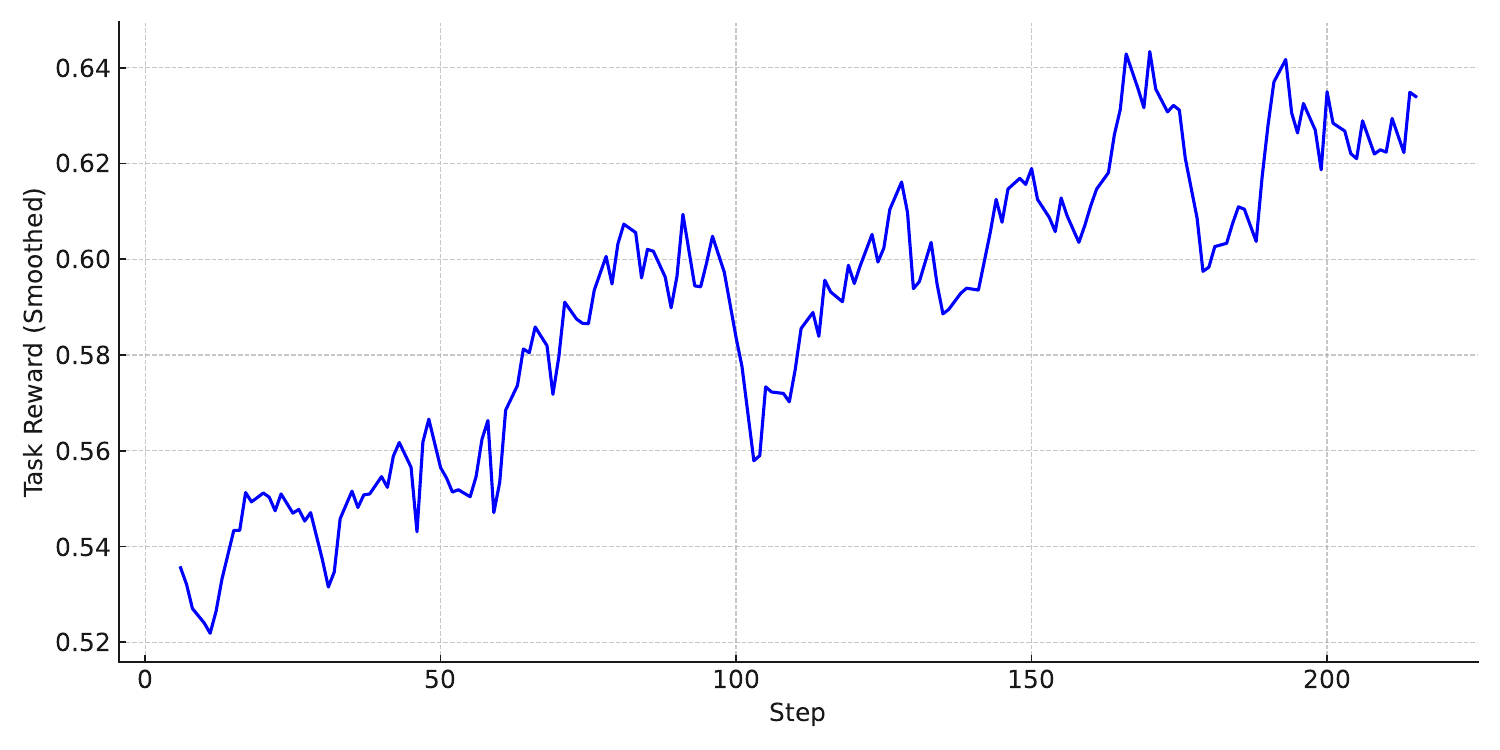}
    \caption{Task Rewards for \shortexperiment}
  \end{subfigure}
  \hfill
  \begin{subfigure}{0.45\textwidth}
    \includegraphics[width=\linewidth]{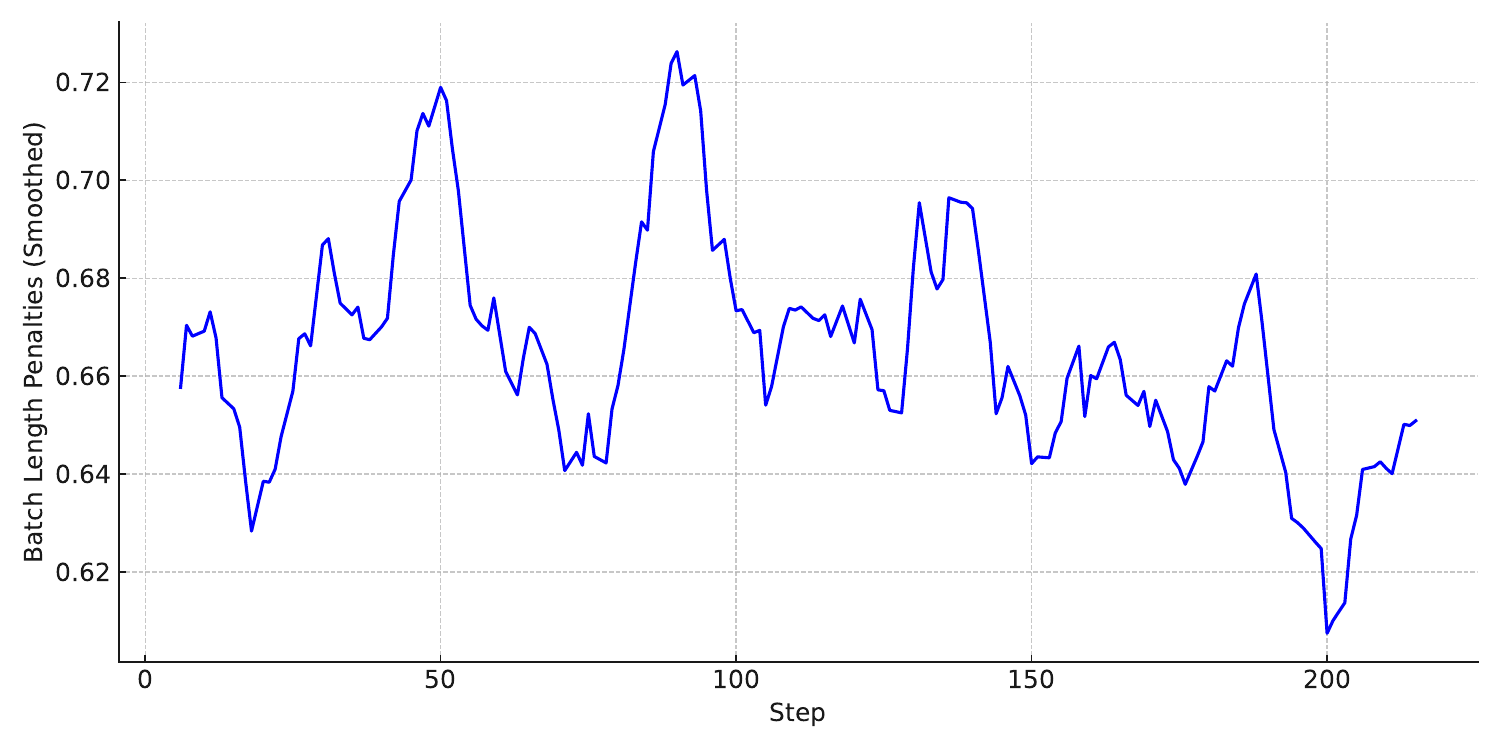}
    \caption{Length penalties for \shortexperiment}
  \end{subfigure}

  \vspace{0.5cm}

  \begin{subfigure}{0.45\textwidth}
    \includegraphics[width=\linewidth]{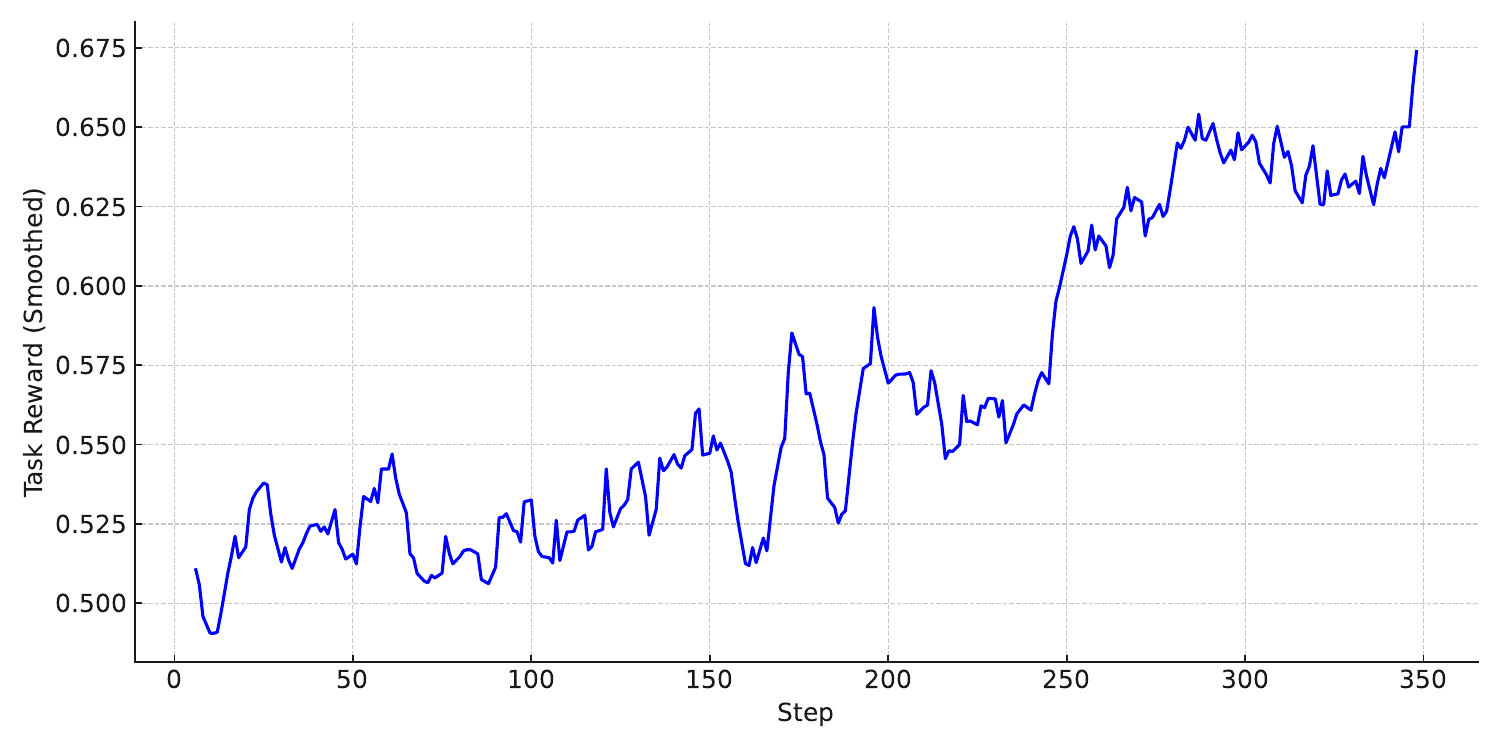}
    \caption{Task Rewards for \longexperiment}
  \end{subfigure}
  \hfill
  \begin{subfigure}{0.45\textwidth}
    \includegraphics[width=\linewidth]{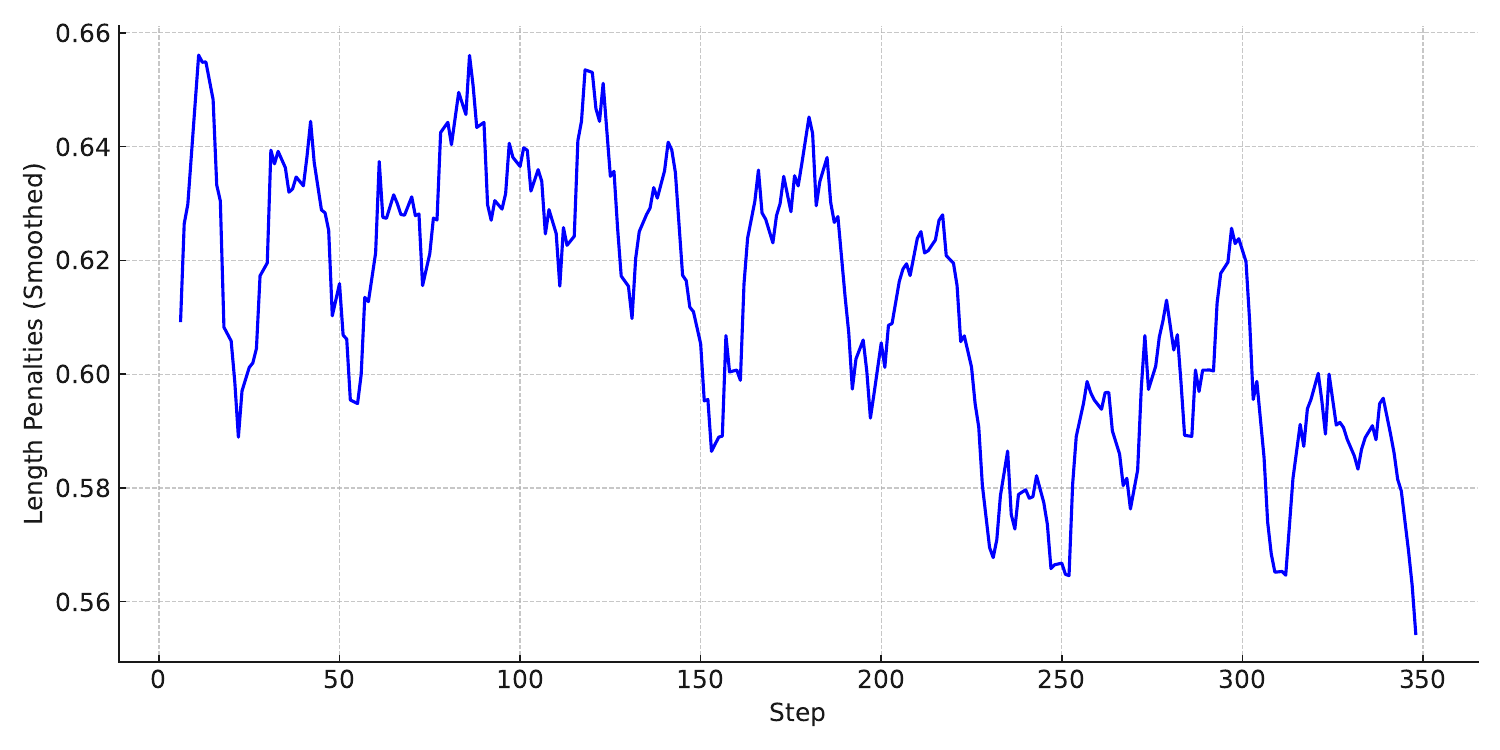}
    \caption{Length penalties for \longexperiment}
  \end{subfigure}
    \caption{Trajectories (smoothed by 10-step moving average) of task rewards and length penalties for \shortexperiment, \ our training run with target lengths \(\{1000, 2000, 3000, 4000\}\) and \longexperiment, \ our training run with target lengths \(\{2000, 4000, 6000, 8000, 10000\}\). During both runs, our task rewards, indicating the ability to solve mathematics and coding problems, rose significantly. In contrast to our ablation experiments with smaller models, length penalties decreased significantly slower; while a clear downward trend is recognizable, the training runs were too short for the length penalties to converge and equip the model with the ability to precisely adhere to a thinking budget.}
    \label{fig:rewards-int2}
\end{figure*}

\paragraph{Reward Trajectories} Throughout training, we saw significant improvements of our task rewards, indicating that the model improved its performance on our mathematics and coding problems. We also saw a reduction of length penalties, but a much slower one than during our ablation experiments with 1.5B and 7B parameter models. As a result, our models did not learn to strictly adhere to the specified thinking budget within the given timeframe of our experiments. Further exploration will be needed to determine what exactly the cause of this is and whether a stronger weighting of the length rewards could have helped learn the length following objective. The full reward trajectories can be found in \autoref{fig:rewards-int2}

\paragraph{Benchmark Performance} We use evalchemy\footnote{https://github.com/mlfoundations/Evalchemy} \citep{evalchemy} along with its default settings to test the performance of our model on common reasoning benchmarks. We use the provided default prompts for Deepseek-R1, QwQ-32B and DeepSeek-R1-Distill-Qwen-32B and attach the length control prompt "Think for \(l_{\mathrm{target}}\) tokens before giving a response" to \intellect. Since the length penalty was not affected significantly during training, we only evaluate our model with the longest target length of 10,000.

As can be seen in \autoref{tab:model_comparison}, we were able to increase the performance of QwQ-32B on mathematics and coding benchmarks, while seeing a slight drop on IFEval, which is likely caused by us solely training on mathematics and coding tasks rather than using more general instruction-following tasks. Overall, as QwQ-32B was already extensively trained with reinforcement learning, it was difficult to obtain huge amounts of generalized improvement on benchmarks beyond our improvements on the training dataset. To see stronger improvements, it is likely that better base models such as the now available Qwen3 \citep{qwen3}, or higher quality datasets and RL environments are needed.

\begin{table*}
\centering
\setlength{\tabcolsep}{4pt}
\begin{tabular}{lccccc}
\toprule
\textbf{Model}  & \textbf{AIME24} & \textbf{AIME25} & \textbf{LiveCodeBench (v5)} & \textbf{GPQA-Diamond} & \textbf{IFEval} \\
\midrule
\intellect         & \textbf{78.8}                & 64.9                & \textbf{67.8}                & 66.8                & 81.5                \\
QwQ-32B         & 76.6                & 64.8                & 66.1                & 66.3                & 83.4                \\
Qwen-R1-Distill-32B         & 69.9                & 58.4                & 55.1                & 65.2               & 72.0                \\
Deepseek-R1         & 78.6                & 65.1                & 64.1                & 71.6                & 82.7                \\
\bottomrule
\end{tabular}
\caption{Performance comparison of models across benchmarks.}
\label{tab:model_comparison}
\end{table*}

\section{Discussion: Decentralized Training in the Test-Time-Compute Paradigm}
\label{sec:discussion-decentralized-ai}

As the compute demands of large language models have increased by several orders of magnitude in recent years, distributed training across data centers has become increasingly relevant. Beyond offering an economically sustainable path for collaborative open-source development, the sheer compute power and energy required to train these models will soon outpace even the largest data centers in the world.

So far, most progress has come from scaling parameters and dataset size—commonly referred to as pretraining scaling. More recently, a complementary axis of progress has emerged: test-time compute scaling, as seen in reasoning-focused models.

While both forms of scaling are compatible with decentralization, we argue that test-time compute scaling is particularly well-suited for decentralized training. It reduces coordination requirements and shifts compute demands toward inference, enabling broader participation from heterogeneous devices.

\paragraph{Asynchronous RL Hides Most Communication Overhead}

Communication is the primary bottleneck in decentralized training. Techniques such as DiLoCo \citep{diloco} can reduce pre-training communication overhead by up to two orders of magnitude. However, as model sizes increase, communication—especially blocking communication—once again becomes the limiting factor.

A promising strategy is to overlap communication with computation. Unlike approaches such as ZeRO-offload \citep{Zerocpu}, which delay gradient application and impact convergence, we argue that delaying rollouts in RL yields a better tradeoff. This is because the delay applies at the model level, not the optimization step. Even if the model is slightly off-policy, it can still generate useful reasoning traces that lead to positive rewards, which are valid training signals.

Further investigation is needed to evaluate asynchronous RL with delays beyond two steps. Nonetheless, with delays of 4–5 steps, we could effectively hide various blocking stages in the RL pipeline—including weight broadcasting, environment verification, permissionless validation, and relative KL log-probability computation. This strategy improves compute utilization across both training and inference and enables greater hardware heterogeneity. Slower devices can still contribute valuable samples. Additionally, decentralized pipeline-parallel inference can facilitate the use of large models on consumer-grade hardware.

\paragraph{Inference Will Consume the Majority of Compute}

In \intellect, the training-to-inference compute ratio was approximately 1:4. We anticipate this ratio will shift even more heavily toward inference as test-time reasoning scales. This trend opens the door to training models with hundreds of billions of parameters on globally distributed heterogeneous compute resources.

A key driver of this shift is dataset filtering. As illustrated in \autoref{fig:difficulty_filtering}, model capabilities improve when training focuses on more challenging samples. However, not all data generated during inference is useful. As models tackle harder tasks with increasingly sparse positive rewards, inference will demand substantially more compute than training. In this setting, generating high-quality rollouts becomes the dominant compute cost. Since only a small subset of these rollouts contains strong learning signals, the majority of compute is allocated to exploration rather than model updates.

This asymmetry in compute demand reshapes the scaling dynamics of decentralized RL and indirectly addresses one of its historical limitations: memory constraints. By shifting most of the workload to inference—where memory requirements are significantly lower than during training—decentralized training becomes feasible at scale across a broader range of hardware.

\section{Conclusion \& Future Work}
\label{sec:conclusion}

In this report, we introduce \intellect, the first globally distributed reinforcement learning run of a 32-billion-parameter language model. We are open-sourcing the trained model, tasks \& verifier environments along with all infrastructure components including our training framework \primeframework. We hope that this report and the accompanying open-source components will support the broader research community in exploring decentralized training, and help advance globally distributed reinforcement learning as a foundation for building frontier open-source models.

While \intellect \ is a first step towards open frontier reasoning models trained in a decentralized fashion, several avenues for future work remain open:

\paragraph{Increasing the Ratio of Inference to Training Compute in Reinforcement Learning} As inference is infinitely parallelizable and does not require any communication between workers, RL training recipes that spend higher amounts of compute on inference relative to training are ideally suited for decentralized training. Methods such as VinePPO \citep{kazemnejad2024vineppounlockingrlpotential} spend additional time on inference to compute Monte Carlo-based value estimates rather than leveraging a value network such as PPO, and are thus an interesting field of study to explore. Additionally, various forms of online data filtering for curriculum learning approaches are purely based on inference, and are thus favorable for decentralized setups, if proven effective.

\paragraph{Tool Calls for Reasoning Models} The latest generation of proprietary reasoning models have access to tool calls such as web browsing, code interpreters and APIs as part of their reasoning chain. Initial promising research results in this direction have come from various open source research efforts \citep{brown2025verifiers, cao2025skyrl, ragen}, opening the door to scaling these methods further and training larger open-source reasoning models capable of leveraging such tools (perhaps even model calls).

\paragraph{Crowdsourcing RL tasks and environments} To teach models new skills, diverse RL environments have to be built. This boils down to a traditional software engineering problem which is highly parallelizable and requires various contributors with specialized areas of domain expertise, making it ideally suited for open-source, community-driven efforts. We invite everyone to contribute RL environments to \primeframework \ and are aiming to make it as easy as possible to crowdsource reinforcement learning environments. 

\paragraph{Model Merging and DiLoCo}
Model merging has emerged as an effective post-training technique in recent work \citep{cohere2025commandaenterprisereadylarge, gemmateam2025gemma3technicalreport, ramé2024warpbenefitsweightaveraged}. Whether such methods extend to reasoning tasks remains an open question. However, the ability to merge models trained on distinct reasoning domains would mark a significant step toward scaling asynchronous reinforcement learning across parallel compute resources. In the merging setup, multiple models could be trained independently and later merged into a single unified model. This could be done at the end of training or continuously during training using techniques like DiLoCo \citep{diloco}, originally developed to reduce communication in data-parallel pretraining. Applying merging in RL would enable scaling decentralized training to one more order of magnitude more compute.

\section*{Acknowledgements}

We would like to thank all the compute contributors for this training run. This includes Demeter Compute, string, 
@BioProtocol, @mev\_pete, @plaintext\_cap, @skre\_0, @oldmankotaro, plabs, @ibuyrugs,  @0xfr\_, @marloXBT, @herb0x\_,  mo, @toptickcrypto, cannopo, @samsja19, @jackminong and primeprimeint1234.

We would also like to thank Michael Luo for his advice on replicating the DeepScaler results.

\bibliography{references}{}
\bibliographystyle{plain}

\end{document}